\documentclass[10pt]{article}
\usepackage[letterpaper,margin=1in]{geometry}
\usepackage[T1]{fontenc}
\usepackage[utf8]{inputenc}
\usepackage{lmodern,microtype}
\usepackage{amsthm,amsmath,amsfonts,amssymb,mathtools,bm,mathrsfs}
\usepackage[authoryear,round]{natbib}
\setcitestyle{authoryear,round,citesep={;},aysep={,},yysep={;}}
\usepackage{booktabs,array,enumitem,xcolor,needspace}
\usepackage[font=small]{caption}
\usepackage{xurl}
\usepackage{graphicx,float,placeins,tabulary}
\newcolumntype{Y}{L}
\floatstyle{plaintop}
\restylefloat{table}
\usepackage{hyperref}
\usepackage{bookmark}
\usepackage[nameinlink,capitalise,noabbrev]{cleveref}
\makeatletter
\renewcommand{\section}{\@startsection{section}{1}{\z@}%
  {-9pt plus -2pt minus -1pt}{5pt plus 1pt minus 1pt}%
  {\normalfont\large\bfseries}}
\renewcommand{\subsection}{\@startsection{subsection}{2}{\z@}%
  {-7pt plus -2pt minus -1pt}{4pt plus 1pt minus 1pt}%
  {\normalfont\normalsize\bfseries}}
\renewcommand{\subsubsection}{\@startsection{subsubsection}{3}{\z@}%
  {-6pt plus -2pt minus -1pt}{3pt plus 1pt minus 1pt}%
  {\normalfont\normalsize\bfseries}}
\renewcommand{\paragraph}{\@startsection{paragraph}{4}{\z@}%
  {5pt plus 1pt minus 1pt}{-0.7em}%
  {\normalfont\normalsize\bfseries}}
\makeatother
\AtBeginDocument{%
  \setlength{\abovedisplayskip}{6pt plus 2pt minus 2pt}%
  \setlength{\belowdisplayskip}{6pt plus 2pt minus 2pt}%
  \setlength{\abovedisplayshortskip}{0pt plus 2pt}%
  \setlength{\belowdisplayshortskip}{4pt plus 2pt minus 2pt}%
}

\allowdisplaybreaks[0]
\newtheoremstyle{preprintplain}
  {5pt plus 1pt minus 1pt}{5pt plus 1pt minus 1pt}
  {\itshape}{}{\bfseries}{.}{0.5em}{}
\newtheoremstyle{preprintdefinition}
  {5pt plus 1pt minus 1pt}{5pt plus 1pt minus 1pt}
  {\normalfont}{}{\bfseries}{.}{0.5em}{}
\theoremstyle{preprintplain}
\newtheorem{theorem}{Theorem}
\newtheorem{corollary}{Corollary}
\newtheorem{proposition}{Proposition}

\theoremstyle{preprintdefinition}

\newenvironment{keywords}
  {\par\smallskip\begingroup\small\noindent\textbf{Keywords.} }
  {\par\endgroup\medskip}

\crefname{lemma}{Lemma}{Lemmas}
\Crefname{lemma}{Lemma}{Lemmas}

\crefname{assumption}{Assumption}{Assumptions}
\Crefname{assumption}{Assumption}{Assumptions}

\crefname{theorem}{Theorem}{Theorems}
\Crefname{theorem}{Theorem}{Theorems}
\crefname{corollary}{Corollary}{Corollaries}
\Crefname{corollary}{Corollary}{Corollaries}
\crefname{proposition}{Proposition}{Propositions}
\Crefname{proposition}{Proposition}{Propositions}
\crefname{definition}{Definition}{Definitions}
\Crefname{definition}{Definition}{Definitions}
\crefname{remark}{Remark}{Remarks}
\Crefname{remark}{Remark}{Remarks}
\crefname{equation}{Equation}{Equations}
\Crefname{equation}{Equation}{Equations}
\crefname{section}{Section}{Sections}
\Crefname{section}{Section}{Sections}
\crefname{figure}{Figure}{Figures}
\Crefname{figure}{Figure}{Figures}
\crefname{table}{Table}{Tables}
\Crefname{table}{Table}{Tables}

\newcommand{\Task}{T}
\newcommand{\bn}{\mathbf n}
\newcommand{\Dn}{D_{\bn}}
\newcommand{\KT}{\mathsf K_{\theta}}
\newcommand{\Ftg}{F_{\theta,\mathbf g}}
\newcommand{\bg}{\mathbf g}
\newcommand{\Gg}{G_{\bg}}
\newcommand{\mJ}{\mathcal J_{\bn}}
\newcommand{\mBjoint}{\mathfrak B_{\bn}}
\newcommand{\taskR}{\widehat{\mathfrak R}_{\bn}}
\newcommand{\popR}{\mathcal R}
\newcommand{\empR}{\widehat{\mathcal R}_{\bn}}
\newcommand{\HS}{\mathrm{HS}}
\newcommand{\opn}{\mathrm{op}}
\newcommand{\loss}{\ell}
\newcommand{\Send}{S_{\theta,\bg}^{\mathrm{end}}}
\newcommand{\Ojoint}{\Omega_{\mathrm{joint}}}
\newcommand{\Osep}{\Omega_{\mathrm{sep}}}
\newcommand{\Jphi}{J_{\Phi}}
\newcommand{\mphi}{m_{\Phi}}
\newcommand{\Mphi}{M_{\Phi}}

\hypersetup{
  colorlinks=true,
  linkcolor=blue!50!black,
  citecolor=green!35!black,
  urlcolor=blue!55!black,
  pdftitle={Trace-Norm Geometry of Composed Multitask Predictors},
  pdfauthor={Mahdi Mohammadigohari},
  pdfsubject={Trace-norm geometry and regularization of composed multitask predictors},
  pdfkeywords={Multitask learning; trace-norm regularization; shared representations; unequal sample sizes; composition operators},
  pdfpagemode=UseNone
}
\AtBeginDocument{\raggedbottom}

\title{Trace-Norm Geometry of\\
Composed Multitask Predictors}
\author{%
\begin{minipage}[t]{0.90\textwidth}\centering\small
Mahdi Mohammadigohari\\
Faculty of Engineering\\
Free University of Bozen--Bolzano\\
\texttt{mahdi.mohammadigohari@gmail.com}
\end{minipage}}
\date{}

\begin{document}
\pdfbookmark[0]{Main paper}{main-paper}
\maketitle

\begin{abstract}
Multitask predictors apply task-specific functions to a shared representation. Penalizing the representation and task functions separately can assign different costs to equivalent predictors and charge for directions the tasks do not use. We study the trace norm of the complete task predictors in a fixed input reproducing kernel Hilbert space, weighted by task sample sizes. For a finite-dimensional intermediate function space, this norm is the infimum of the product of the representation operator norm and task trace norm over invertible coordinate changes. Task inner products determine this norm; standard duality gives the exact empirical Rademacher complexity of its norm ball. We estimate the feature Gram matrix used in fitting. Under bounded leverage, one estimate controls the penalty for every task head and shared linear combination of fixed features, even when the Gram is singular and the coefficients are chosen after estimation. This yields comparisons of penalized empirical objectives and a generalization bound under an estimated norm constraint, with the stated loss and sampling assumptions. Controlled experiments show that signal alignment can favor either penalty. A SARCOS comparison finds close agreement between exact and estimated penalties on overlapping training and evaluation observations.
\end{abstract}


\section{Introduction}\label{sec:intro}

Multitask learning shares features across prediction problems with limited or unequal data~\citep{Evgeniou2005,Argyriou2008convex,Pontil2013Excess,Maurer2016Benefit}. For $\Task$ tasks, let $\Phi_\theta:\mathcal X\to\mathcal Z$ be a shared network with parameters $\theta$, mapping inputs to features. Task $t$ uses an output function $g_t:\mathcal Z\to\mathbb R$, or task head, to predict $f_t(x)=g_t(\Phi_\theta(x))$. Different intermediate coordinates can represent the same predictors, yet separate penalties on the representation and task heads may assign them different costs. Such penalties can also charge for directions that the tasks do not use. We ask how to regularize the complete predictors directly, and how to do so when their inner products must be estimated.

Let $\mathcal H_L$ be a Hilbert space of task heads on $\mathcal Z$ and $\mathcal H_0$ a fixed reproducing kernel Hilbert space (RKHS) on $\mathcal X$. For fixed $\theta$, the network induces the composition operator $\KT:g\mapsto g\circ\Phi_\theta$, where $g\in\mathcal H_L$ is a generic task head. We assume that $\KT:\mathcal H_L\to\mathcal H_0$ is bounded. It is linear in $g$, even when the network is nonlinear in $x$. Composition operators are also called Koopman operators; \citet{hashimoto2024koopmanbased} use them to analyze neural-network generalization. Collect the task heads $\bg=(g_1,\ldots,g_\Task)$ in $\Gg$ and their predictors in $\Ftg$:
\begin{align}
\Gg c=\sum_{t=1}^{\Task}c_tg_t,
\qquad c\in\mathbb R^{\Task},
\label{eq:task-map-intro}\\
\Ftg=\KT\Gg:\mathbb R^{\Task}\to\mathcal H_0,
\qquad \Ftg e_t=f_t.
\label{eq:end-to-end-map}
\end{align}
Here $e_t$ is the $t$th standard basis vector. Let $\bn=(n_1,\ldots,n_\Task)$, where $n_t>0$ is the number of observations for task $t$, and write $\Dn=\operatorname{diag}(n_1,\ldots,n_\Task)$. For a predictor map $F$, define
\begin{align}
\mJ(F)=\lVert F\Dn^{-1/2}\rVert_*.
\label{eq:joint-geometry}
\end{align}
The trace (nuclear) norm $\lVert\cdot\rVert_*$ sums the singular values and measures the task functions jointly. The inverse-square-root task weights are fixed before sampling~\citep{Pontil2013Excess}. The usual separated bound is
\begin{align}
\mJ(\KT\Gg)\le\lVert\KT\rVert_{\opn}
\lVert\Gg\Dn^{-1/2}\rVert_*
\label{eq:intro-separated-upper}
\end{align}
The operator norm uses the largest amplification of $\KT$ over all unit-norm functions in $\mathcal H_L$, including those unused by the tasks. This can make the bound arbitrarily larger than the norm of the actual predictors. Taking the norm of $\KT\Gg$ retains how the shared network acts on the chosen task heads.

Our main characterization identifies the infimum of these bounds. For bounded linear maps $K:\mathcal H_L\to\mathcal H_0$ and $G:\mathbb R^{\Task}\to\mathcal H_L$, with $\mathcal H_L$ finite dimensional,
\begin{align*}
\mJ(KG)=\inf_{R\in\operatorname{GL}(\mathcal H_L)}
\lVert KR^{-1}\rVert_{\opn}\,
\lVert RG\Dn^{-1/2}\rVert_*.
\end{align*}
Here $R$ ranges over invertible linear maps on $\mathcal H_L$; \Cref{thm:optimal-refactor} also covers non-injective $K$. The coordinate changes act on the intermediate function space and need not all be realizable by a prescribed neural architecture. The task Gram is
\begin{align}
\Send=\Ftg^*\Ftg=\Gg^*\KT^*\KT\Gg.
\label{eq:end-to-end-gram}
\end{align}
Here $^{*}$ denotes the Hilbert-space adjoint, and the entries of $\Send$ are the input-RKHS inner products of the task predictors. The eigenvalues of $\Dn^{-1/2}\Send\Dn^{-1/2}$ determine their weighted trace norm and its equivalent description through task covariances. Standard norm duality gives the exact complexity of the class with a fixed bound on this norm; any contained neural class inherits an upper bound.

For fixed nonlinear features $b_1,\ldots,b_p\in\mathcal H_0$, write $f_t=\sum_{j=1}^p A_{jt}b_j$, with $A\in\mathbb R^{p\times\Task}$, and let $Q_{ij}=\langle b_i,b_j\rangle_{\mathcal H_0}$. The penalty is $\|Q^{1/2}A\Dn^{-1/2}\|_*$. A relative approximation of $Q$ preserves squared norms throughout the feature span up to the same relative error, and controls the penalty for every $A$. Under bounded leverage, sufficiently many independent reference inputs give this approximation, even for singular $Q$ and shared linear combinations fitted after estimation. This gives comparisons of penalized empirical objectives and, under an estimated norm constraint, a generalization bound.

\paragraph{Contributions.}
\begin{itemize}[leftmargin=1.35em,itemsep=.18em,topsep=.2em]
\item \textbf{Geometry of composed multitask predictors.}
We characterize the best separated bound over intermediate coordinates and quantify the effects of task alignment and cancellation.
\item \textbf{Statistical control of composed predictors.}
Standard trace/operator duality gives the exact complexity of a fixed predictor ball; an example quantifies the Hilbert--Schmidt relaxation gap.
\item \textbf{Regularization with an estimated Gram matrix.}
Uniform Gram approximation yields fitting and generalization guarantees for task heads and shared linear maps selected within a fixed feature span.
\end{itemize}

\Cref{sec:notation,sec:setting} give notation and the sampling assumptions. The geometry, complexity, estimation and experimental results follow in \Cref{sec:composed-geometry,sec:exact-complexity,sec:estimated-feature-geometry,sec:experiments}; proofs and protocols are in the appendices.

\section{Related work}\label{sec:related}

\paragraph{Trace-norm multitask learning.}
Kernel methods encode relations between tasks through matrix-valued kernels~\citep{Evgeniou2005,Micchelli2005,Caponnetto2008}. Learning a matrix of task relations connects shared feature learning to trace-norm regularization~\citep{Argyriou2008convex,dinuzzo2011lowrank,Ciliberto2015ConvexTasks}. Our weighting already appears in \citet[Appendix~C]{Pontil2013Excess}: with $W=F^*$ and $S=\operatorname{diag}(\sqrt{\bar n/n_t})$, where $\bar n=\Task^{-1}\sum_t n_t$, one has $\lVert SW\rVert_* = \sqrt{\bar n}\,\mJ(F)$. Related weights in matrix completion depend on row and column sampling probabilities~\citep{SalakhutdinovSrebro2010Weighted,FoygelEtAl2011Weighted}. Generalization analyses cover shared feature learning~\citep{Maurer2014Structured,Maurer2016Benefit}, local Rademacher complexity~\citep{Yousefi2018Local} and few observations per task~\citep{Boursier2022Scarce,Liu2023ImprovedTrace}. We apply the established trace norm to task functions after the shared representation and study its estimation from feature inner products.

\paragraph{Composition operators and neural-network bounds.}
\citet{hashimoto2024koopmanbased} express scalar-output neural networks as successive composition operators and use their norms to bound Rademacher complexity. Their Sobolev results cover invertible and injective rectangular weights. For non-injective weights, they construct modified predictors that agree with the original networks on suitable compact regions. They also explain how bounding each layer separately can lose information about the combined action of the layers.

\citet{MohammadigohariBorsaniDiFatta2026Operator} extend this analysis to multiple outputs with a fixed matrix describing task relations. They also study learning a shared linear map between function spaces, with fixed task functions and a squared Hilbert--Schmidt penalty on the shared map. We study the weighted trace norm of the complete predictor map $F=KG$. Its task Gram matrix, $G^*K^*KG$, records the inner products of the task predictors after composition. Replacing $K^*K$ by the upper bound $\|K\|_{\opn}^2I$ gives the separated norm bound. Our coordinate-change theorem assumes that $K$ is bounded between the chosen function spaces and allows $K$ to be non-injective.

\paragraph{Matrix factorizations and network regularization.}
Trace norms and Schatten quasi-norms can be expressed by minimizing penalties over all compatible matrix factorizations~\citep{ShangLiuCheng2016Tractable,ShangLiuCheng2016Unified}. Our result starts from a given factorization $F=KG$ and optimizes the product of the operator and trace norms over invertible coordinate changes $(KR^{-1},RG)$. A related literature asks which penalties on represented functions are induced by penalties on network parameters~\citep{SavareseEtAl2019FunctionSpace,EMaWu2022BarronFlow,OngieParhi2026RepresentationCosts}. This includes the effect of adding linear layers to ReLU networks~\citep{OngieWillett2022LinearLayers} and the relation between weight decay and shared features in multi-output networks~\citep{ShenoudaEtAl2024Variation}.

\paragraph{Approximating Gram matrices.}
Sampled Gram matrices support stability guarantees for weighted least squares~\citep{CohenMigliorati2017}, while covariance approximation supports statistical guarantees for Nystr\"om kernel ridge regression~\citep{RudiCamorianoRosasco2015}. \citet[Section~5, Remark~43 and Lemma~49]{AvronClarksonWoodruff2017Sharper} show that approximately preserving Euclidean norms on a fixed subspace also approximates nuclear norms for every coefficient matrix. We apply this principle to the weighted trace penalty, estimating the feature Gram from independent inputs by matrix Chernoff concentration~\citep[Corollary~5.2]{Tropp2012UserFriendly}. Under bounded leverage, the bound holds for singular Grams and all fitted linear combinations of the fixed features. For inner products defined through derivatives, the chain rule and area formula~\citep{EvansGariepy2015} give an integral formula that sums contributions from all preimages of a finite-to-one map. This extends the Cameron--Martin calculation for diffeomorphic activations in \citet{MohammadigohariBorsaniDiFatta2026Operator}; \Cref{sec:pullback} states the assumptions.

\section{Notation}\label{sec:notation}

For a bounded linear map $A$, $A^*$ denotes its Hilbert-space adjoint, and $\ker A$ and $\operatorname{range}(A)$ denote its kernel and range. We write $\lVert A\rVert_{\opn}$ for the operator norm, $\lVert A\rVert_{\HS}$ for the Hilbert--Schmidt norm, and $\lVert A\rVert_*$ for the trace (nuclear) norm, when finite. In finite dimensions, the Hilbert--Schmidt norm is the Frobenius norm. The notation $\operatorname{Tr}$ denotes the trace. For symmetric matrices, $A\preceq B$ means that $B-A$ is positive semidefinite; $\mathbb S_+^d$ and $\mathbb S_{++}^d$ denote the positive-semidefinite and positive-definite $d\times d$ matrices. For $A\succeq0$, $A^{1/2}$ is its positive-semidefinite square root and $A^\dagger$ its Moore--Penrose inverse. We write $\operatorname{GL}(\mathcal H)$ for the invertible linear maps on a finite-dimensional Hilbert space $\mathcal H$.

\section{Setting and composed task predictors}\label{sec:setting}

Fix $\Task\ge1$ and positive integer task counts $n_1,\ldots,n_\Task$ before sampling. Within task $t$, the observations $S_t=\{(x_{ti},y_{ti})\}_{i=1}^{n_t}$ are i.i.d. from $P_t$ on the measurable space $\mathcal X\times\mathcal Y_t$, and observations are independent across tasks. The risk weights tasks equally; tasks without observations are excluded. We use $\bn=(n_1,\ldots,n_\Task)$ and $\Dn=\operatorname{diag}(n_1,\ldots,n_\Task)$ as above.

Let $\mathcal H_0$ be a fixed separable real scalar RKHS, chosen independently of the observations, with measurable kernel and predictors on $\mathcal X$. Its kernel $k_0$ satisfies
\begin{align}
\sup_{x\in\mathcal X}k_0(x,x)
\le
\kappa.
\label{eq:kernel-bound}
\end{align}
For a bounded linear map $F:\mathbb R^\Task\to\mathcal H_0$, define predictors $f_t=Fe_t$ and the trace norm weighted by task sample sizes
\begin{align}
\mJ(F)
=
\left\lVert
F\Dn^{-1/2}
\right\rVert_*.
\label{eq:joint-norm-definition}
\end{align}
Because the domain is finite dimensional, every such $F$ is finite rank and the nuclear norm is well defined.

For $\tau\ge0$, define the trace-norm ball
\begin{align}
\mBjoint(\tau)
=
\left\{
F:\mathbb R^\Task\to\mathcal H_0:
\mJ(F)\le\tau
\right\}.
\label{eq:joint-ball}
\end{align}

When a shared network and task heads are available, $F=\Ftg=\KT\Gg$ as in \Cref{eq:end-to-end-map}.  We first analyze the resulting predictor map, without requiring its factorization to be unique or observed.  This gives the predictor norm against which we compare separate representation and task bounds.

\section{Geometry of composed task predictors}\label{sec:composed-geometry}

Changing intermediate coordinates preserves $KG$ and its weighted trace norm, but can change the separated product. Minimizing over these coordinates characterizes the composed norm. Optimization over task covariances gives a second description through the Gram matrix of the composed task functions.

\subsection{Equivalent intermediate coordinates}\label{sec:refactorization}

\begin{theorem}[Best separated bound over equivalent intermediate coordinates]\label{thm:optimal-refactor}
Assume that $\mathcal H_L$ is finite dimensional.  For every bounded
$K:\mathcal H_L\to\mathcal H_0$ and every
$G:\mathbb R^\Task\to\mathcal H_L$,
\begin{align}
\inf_{R\in\operatorname{GL}(\mathcal H_L)}
\left\lVert
KR^{-1}
\right\rVert_{\opn}
\left\lVert
RG\Dn^{-1/2}
\right\rVert_*
=
\left\lVert
KG\Dn^{-1/2}
\right\rVert_*.
\label{eq:optimal-refactor}
\end{align}
If $K$ is injective, the infimum is attained by
$R=\left(K^*K\right)^{1/2}$, up to a positive scalar.  For singular $K$, the proof gives an explicit invertible sequence that suppresses task components in $\ker K$ and approaches the same value.
\end{theorem}

The proof, including the singular construction, is in \Cref{app:proof-coordinate}.

For injective $K$, the transformed map $KR^{-1}$ preserves norms when $R=(K^*K)^{1/2}$, so the separated bound is exact. For singular $K$, the approximating sequence reduces the contribution of task components that composition removes.

\paragraph{General coordinate changes versus changes allowed by an architecture.}
The infimum in \Cref{thm:optimal-refactor} ranges over all $R\in\operatorname{GL}(\mathcal H_L)$.  A fixed neural architecture or a prescribed RKHS may realize only a subset $\mathfrak R_{\mathrm{arch}}$ of these maps, in which case
\begin{align*}
\inf_{R\in\mathfrak R_{\mathrm{arch}}}
\lVert KR^{-1}\rVert_{\opn}
\lVert RG\Dn^{-1/2}\rVert_*
\ge
\lVert KG\Dn^{-1/2}\rVert_*.
\end{align*}
Admitting the balancing coordinates, or the approximating sequence from the proof, is sufficient for equality. Other permitted coordinates may also give equality for a particular task map $G$. A realization using linear functions is given in \Cref{app:linear-coordinate-realization}.

\paragraph{How much can separation cost?}
The constructions in \Cref{sec:no-go} show three ways in which separate bounds lose information. Equivalent coordinates can keep the composed norm equal to one while making the separated product $a^2$, for any $a\ge1$. Even with fixed singular values for both factors, changing their relative directions changes the composed norm. Finally, $m$ pairs of cancelling linear layers can represent the identity while their operator norms multiply to $a^{2m}$. These examples explain why measuring the composed predictor can give a smaller radius in complexity bounds. The task Gram below also makes that geometry available for regularization.

\subsection{Task covariance and rank}\label{sec:covariance-summary}

For a fixed predictor map $F$, let $S_F=F^*F$ be its task Gram matrix.
The covariance formulation of output-kernel learning gives
\begin{align}
\mJ(F)
&=\inf_{\Sigma\in\mathbb S_{++}^{\Task}}
\left[\operatorname{Tr}(\Dn^{-1}\Sigma)
      \operatorname{Tr}(\Sigma^{-1}S_F)\right]^{1/2}
\notag\\
&=\operatorname{Tr}\!\left[
  (\Dn^{-1/2}S_F\Dn^{-1/2})^{1/2}\right].
\label{eq:covariance-main}
\end{align}
For $F=\KT\Gg$, the matrix $S_F=\Gg^*\KT^*\KT\Gg$ records the task inner products after composition. \Cref{thm:end-covariance} gives the covariance optimizer and attainment cases, with proof in \Cref{app:proof-covariance}. \Cref{sec:joint-covariance} also gives the associated rank bound and the quotient space that identifies task heads with the same predictor.

\section{Statistical control in a fixed input RKHS}\label{sec:exact-complexity}

Rademacher complexity links a class of predictors to uniform risk bounds~\citep{BartlettMendelson2002}. Multi-output analyses also exploit the coupling among task predictors~\citep{Reeve2020Multioutput}. For the ball $\mBjoint(\tau)$, duality between trace and operator norms gives its empirical complexity directly.

For a fixed input sample, let $\{\varepsilon_{ti}\}$ be independent Rademacher variables and define
\begin{align}
z_t
&=
\frac{1}{n_t}
\sum_{i=1}^{n_t}
\varepsilon_{ti}k_0\left(\cdot,x_{ti}\right),
\qquad
Z_\varepsilon e_t
=
z_t.
\label{eq:sample-operator}
\end{align}
For a class $\mathcal F$ of predictor maps, its empirical Rademacher complexity, with equal weight for each task, is
\begin{align}
\taskR\left(\mathcal F\right)
=
\mathbb E_\varepsilon
\sup_{F\in\mathcal F}
\frac{1}{\Task}
\sum_{t=1}^{\Task}
\frac{1}{n_t}
\sum_{i=1}^{n_t}
\varepsilon_{ti}
\left(Fe_t\right)\left(x_{ti}\right).
\label{eq:rademacher-definition}
\end{align}
\begin{theorem}[Exact empirical complexity of the end-to-end trace-norm class]\label{thm:exact-rademacher}
For every fixed input sample and every $\tau\ge0$,
\begin{align}
\taskR\left(\mBjoint(\tau)\right)
=
\frac{\tau}{\Task}
\mathbb E_\varepsilon
\left\lVert
Z_\varepsilon\Dn^{1/2}
\right\rVert_{\opn}.
\label{eq:exact-rademacher}
\end{align}
Under \Cref{eq:kernel-bound},
\begin{align}
\taskR\left(\mBjoint(\tau)\right)
\le
\tau
\sqrt{\frac{\kappa}{\Task}}.
\label{eq:hs-relaxation}
\end{align}
The identity in \Cref{eq:exact-rademacher} is conditional on the fixed input sample and the fixed class.  Hence the class may contain estimators fitted on that same sample, provided that its radius $\tau$ is specified before the sample is observed.
\end{theorem}

The proof is in \Cref{app:proof-complexity}.

The duality between trace and operator norms~\citep{Simon2005TraceIdeals} pairs $F$ with the random operator $Z_\varepsilon$.  The simpler bound in \Cref{eq:hs-relaxation} uses $\lVert\cdot\rVert_{\opn}\le\lVert\cdot\rVert_{\HS}$ to replace that operator norm by its Hilbert--Schmidt norm.  When the kernel functions $k_0(\cdot,x_{ti})$ are pairwise orthogonal with squared norm $\kappa$, the exact complexity is $\tau\sqrt{\kappa}/\Task$, so the relaxation can lose a factor $\sqrt{\Task}$ (\Cref{cor:orthogonal-sharpness}).

\paragraph{Sample-size interpretation.}
For balanced counts $n_t=n$, the normalization is $\mJ(F)=\lVert F\rVert_*/\sqrt n$. The conventional ball $\lVert F\rVert_*\le B$ therefore corresponds to $\tau=B/\sqrt n$, and \Cref{eq:hs-relaxation} gives $B\sqrt{\kappa/(n\Task)}$. Holding the normalized radius $\tau$ fixed as $n$ increases instead enlarges the unnormalized class to $\lVert F\rVert_*\le\tau\sqrt n$. Comparisons across sample sizes must account for this change of class. The regularized estimator below provides one way to control the radius; localization gives another~\citep{Yousefi2018Local}.

\paragraph{Regularization and the predictor radius.}
Let $\widehat L(F)\ge0$ be an empirical risk and fix $\lambda>0$ before observing the training sample.  Every minimizer $\widehat F$ of $\widehat L(F)+\lambda\mJ(F)$ over $F$ satisfies
\begin{align}
\mJ(\widehat F)
\le
\frac{\widehat L(0)}{\lambda}.
\label{eq:penalty-radius}
\end{align}
Indeed, compare the objective at $\widehat F$ with its value at $F=0$ and use nonnegativity of the loss.  If $\widehat L(0)\le C_0$ almost surely for a deterministic $C_0$, then \Cref{thm:exact-rademacher} applies with the prespecified radius $C_0/\lambda$.  For task-balanced binary logistic loss with zero logits, $C_0=\log 2$.  If $\lambda$ is chosen from the data but has a deterministic lower bound $\lambda_{\min}>0$, the same argument places every selected minimizer in the prespecified ball of radius $C_0/\lambda_{\min}$. Alternatively, one may use a risk bound that holds simultaneously over the candidate radii.

For any composed class contained in $\mBjoint(\tau)$,
\Cref{thm:exact-rademacher} gives a uniform upper bound.
Separate constraints $\lVert\KT\rVert_{\opn}\le\Gamma$ and
$\lVert\Gg\Dn^{-1/2}\rVert_*\le\rho$ suffice for containment with
$\tau=\Gamma\rho$; \Cref{cor:separated-relaxation} gives the complete
complexity bound and proof in \Cref{app:proof-separated}.
For bounded Lipschitz losses, \Cref{app:population} converts the complexity
identity into a uniform population-risk bound.
\Cref{sec:estimated-generalization} states the consequence when the norm
itself is estimated.

\section{Estimating the feature Gram matrix}
\label{sec:estimated-feature-geometry}

For a fixed finite family of shared features, one approximation of their
Gram matrix controls the regularizer for every fitted task head. We first
establish this approximation, then use it for fitting and generalization.

\subsection{Uniform approximation in a fixed feature span}\label{sec:uniform-feature-geometry}
Let $b_1,\ldots,b_p\in\mathcal H_0$, define
$K:\mathbb R^p\to\mathcal H_0$ by $Ke_j=b_j$, and set
$Q=K^*K\in\mathbb S_+^p$.  For a matrix of task coefficients
$A\in\mathbb R^{p\times\Task}$, the composed predictor map is $F_A=KA$.
Its regularizer is
\begin{align}
J_Q(A)
&:=\lVert Q^{1/2}A\Dn^{-1/2}\rVert_*
\notag\\
&=\operatorname{Tr}\!\left[
 \left(\Dn^{-1/2}A^\top Q A\Dn^{-1/2}\right)^{1/2}
 \right]
=\mJ(F_A).
\label{eq:finite-feature-penalty}
\end{align}
The equality with $\mJ(F_A)$ holds because $Q^{1/2}A$ and $KA$ have the same task Gram matrix.  This formula permits redundant features: $Q$ need not be
invertible.  On coefficient matrices, $J_Q$ is a seminorm when $Q$ is
singular; on the represented predictor maps it is the trace norm already
defined in \Cref{eq:joint-ball}.

Suppose this feature Gram admits a sampling representation
\begin{align}
Q=\mathbb E_{U\sim\mu}W(U),
\qquad
\widehat Q=\frac1m\sum_{j=1}^m W(U_j),
\qquad
\widehat J(A)=J_{\widehat Q}(A),
\label{eq:sampled-feature-gram}
\end{align}
where $W$ is a measurable positive-semidefinite $p\times p$ matrix with
finite expected trace.  The feature family and sampling law are fixed
independently of the reference sample $U_1,\ldots,U_m$.

\Needspace{7\baselineskip}
\begin{theorem}[Uniform approximation of the composed regularizer]
\label{thm:estimated-feature-geometry}
Let $U_1,\ldots,U_m$ be independent with law $\mu$, let
$r=\operatorname{rank}(Q)$, and suppose that, for some finite $L\ge1$,
\begin{align}
0\preceq W(U)\preceq LQ\qquad\text{almost surely}.
\label{eq:feature-leverage-bound}
\end{align}
Fix $0<\varepsilon<1$, $0<\delta<1$, and positive task counts
$n_1,\ldots,n_\Task$.  If $r\ge1$ and
\begin{align}
m\ge\frac{3L}{\varepsilon^2}\log\frac{2r}{\delta},
\label{eq:feature-reference-size}
\end{align}
then, with probability at least $1-\delta$,
\begin{align}
(1-\varepsilon)Q
\preceq\widehat Q
\preceq(1+\varepsilon)Q
\label{eq:feature-spectral-sandwich}
\end{align}
and, on the same event, simultaneously for every
$A\in\mathbb R^{p\times\Task}$,
\begin{align}
\sqrt{1-\varepsilon}\,J_Q(A)
\le\widehat J(A)
\le\sqrt{1+\varepsilon}\,J_Q(A).
\label{eq:feature-penalty-sandwich}
\end{align}
On this event $\ker\widehat Q=\ker Q$.  If $r=0$, then
$W(U)=0$ almost surely, so $\widehat Q=Q=0$ and
$\widehat J(A)=J_Q(A)=0$ for every $A$, almost surely, for every $m\ge1$.
\end{theorem}

The proof is in \Cref{app:estimated-feature-geometry}.

The probability bound is obtained by applying the matrix Chernoff inequality
\citep[Corollary~5.2]{Tropp2012UserFriendly} on $\operatorname{range}(Q)$.
The quantity $L$ bounds the largest contribution of one reference point
relative to the feature Gram matrix $Q$.  Equivalently, it bounds
$\lVert Q^{\dagger/2}W(U)Q^{\dagger/2}\rVert_{\opn}$ almost surely;
the dagger denotes the Moore--Penrose inverse.  The proof shows that $W(U)$ vanishes on $\ker Q$ almost surely, as required for this equivalence.  Finite moments alone
do not imply \Cref{eq:feature-leverage-bound}.

The uniformity in $A$ allows the task heads to be fitted using $\widehat Q$.
For every $\tau\ge0$, \Cref{eq:feature-penalty-sandwich} also gives
\begin{align}
\{A:J_Q(A)\le\tau/\sqrt{1+\varepsilon}\}
&\subseteq\{A:\widehat J(A)\le\tau\}
\notag\\
&\subseteq\{A:J_Q(A)\le\tau/\sqrt{1-\varepsilon}\}.
\label{eq:estimated-ball-containment}
\end{align}
Thus a constraint imposed with the estimated Gram matrix controls the composed
norm, after increasing its radius by the displayed factor.  When $\mathcal H_0$ satisfies
the fixed-RKHS assumptions of \Cref{sec:exact-complexity}, the final set
maps into the corresponding ambient ball $\mBjoint(\tau/\sqrt{1-\varepsilon})$.

\Needspace{7\baselineskip}
\subsection{Fitting task heads and shared linear maps}\label{sec:estimated-fitting}

Write $c_-=\sqrt{1-\varepsilon}$ and $c_+=\sqrt{1+\varepsilon}$.
A shared linear map $M\in\mathbb R^{p\times q}$ and task coefficients
$A\in\mathbb R^{q\times\Task}$ give the predictor $F_{M,A}=KMA$.
Let $\mathcal C$ be a nonempty family of these pairs, allowing $q\ge1$
to vary, and let the empirical loss $\widehat L:\mathcal C\to[0,+\infty]$
be finite at some pair. For fixed $\lambda,\eta\ge0$, define
$\mathcal E_\lambda(M,A)=\widehat L(M,A)+\lambda J_Q(MA)$.
On the event in \Cref{eq:feature-spectral-sandwich}, any pair
$(\widehat M,\widehat A)$ that minimizes
$\widehat L(M,A)+\lambda J_{\widehat Q}(MA)$ to within $\eta$ of its
infimum over $\mathcal C$ satisfies
\begin{align}
\mathcal E_\lambda(\widehat M,\widehat A)
\le\frac{c_+}{c_-}\inf_{(M,A)\in\mathcal C}\mathcal E_\lambda(M,A)
+\frac{\eta}{c_-}.
\label{eq:objective-main}
\end{align}
Fixed-feature task fitting is the case $M=I_p$.
\Cref{cor:estimated-geometry-objective,cor:adaptive-feature-span} give the
complete comparisons, proved in
\Cref{app:proof-objective,app:proof-adaptive-span}.

This comparison concerns the same empirical loss with an estimated or exact
geometry. The admissible pairs may satisfy architectural or rank constraints;
neither convexity nor attainment of the infimum is required. For $\eta=0$,
the premise assumes an actual minimizer. Both fitted matrices, including
the width and rank of $M$, may depend on the reference sample because every predictor $KMA$ uses the same fixed feature span.

The feature family may first be learned from separate data: conditioning
on that data gives the same result whenever the assumptions hold
conditionally. Selecting nonlinear features outside the fixed span using
the reference sample requires an additional uniform approximation argument.
\Cref{app:multiple-dictionaries} gives the extension to finitely many
prespecified dictionaries. For inner products involving derivatives,
\Cref{app:derivative-features} gives the feature contribution $W(x)$ and the
conditions under which the feature span is an RKHS.

\subsection{Generalization under an estimated constraint}\label{sec:estimated-generalization}

For $f_t=Fe_t$, write the task-balanced population and empirical risks as
\begin{align*}
\popR(F)&=\frac1\Task\sum_{t=1}^{\Task}
 \mathbb E_{(X,Y)\sim P_t}\loss_t(f_t(X),Y),
\\
\empR(F)&=\frac1\Task\sum_{t=1}^{\Task}\frac1{n_t}
 \sum_{i=1}^{n_t}\loss_t(f_t(x_{ti}),y_{ti}).
\end{align*}
Assume the fixed separable measurable RKHS and sampling setting of
\Cref{sec:setting}, with jointly measurable losses that are
$L_\loss$-Lipschitz in prediction and bounded in $[0,B_\loss]$.
The feature dictionary and its sampling law are fixed independently of the
reference sample, which is independent of the labelled sample.
Fix $\tau\ge0$, $0<\varepsilon<1$, and
$\delta_{\mathrm g},\delta_{\mathrm r}\in(0,1)$ with
$\delta_{\mathrm g}+\delta_{\mathrm r}<1$ before sampling.
Under the leverage condition \eqref{eq:feature-leverage-bound}, choose
$m$ by \Cref{eq:feature-reference-size} with $\delta=\delta_{\mathrm g}$,
using the rank-zero alternative of \Cref{thm:estimated-feature-geometry}
when appropriate. Then, with probability at least
$1-\delta_{\mathrm g}-\delta_{\mathrm r}$, every fitted pair in this span
satisfying $J_{\widehat Q}(MA)\le\tau$ has $F=KMA$ satisfying
\begin{align}
\popR(F)
&\le\empR(F)
 +\frac{2L_\loss\tau}{\Task\sqrt{1-\varepsilon}}
   \mathbb E_\varepsilon
   \lVert Z_\varepsilon\Dn^{1/2}\rVert_{\opn}
\notag\\
&\quad+3B_\loss
 \left[\frac{\log(2/\delta_{\mathrm r})}{2\Task^2}
              \sum_{t=1}^{\Task}\frac1{n_t}\right]^{1/2}.
\label{eq:estimated-risk-main}
\end{align}
The expectation is over the Rademacher signs in
\Cref{eq:sample-operator}; the selected matrices may depend on both
samples. The estimated constraint places all these predictors in the fixed
ball of radius $\tau/\sqrt{1-\varepsilon}$.
\Cref{cor:estimated-radius-population} is the complete formulation, with
proof in \Cref{app:proof-estimated-radius}.

\section{Empirical evaluation}\label{sec:experiments}

The experiments ask how representation geometry affects regularization, how
closely an estimated Gram matrix reproduces an exact penalty, and how unequal
task counts affect the choice of weights. We address these questions with a
controlled nonlinear study, an independent Gram-estimation study, a
comparison on robot data and linear regression experiments. The empirical comparisons use squared error.
The population guarantee in \Cref{cor:estimated-radius-population} applies to
the bounded Lipschitz losses specified there.

\subsection{How composition affects regularization}\label{sec:composition-pilot}

We test how the geometry of a fixed shared representation affects multitask regularization. This prespecified study uses nonlinear input features and balanced task counts, so the comparison isolates composition from count weighting. Let $\varphi$ contain the eight functions $\sqrt{2}\sin(kx)$ and $\sqrt{2}\cos(kx)$, $k=1,\ldots,4$, orthonormal under the uniform distribution on $[-\pi,\pi]$. Their span, equipped with the coefficient inner product, is a finite-dimensional RKHS. Shared features have coefficient matrix $K$, task functions have coefficients $F=KA$, and their feature Gram is $Q=K^\top K$.

We compare the penalties $\|Q^{1/2}A/\sqrt n\|_*$ and $\|A/\sqrt n\|_*$ with the same squared loss. Here $K$ is diagonal, $\|K\|_{\opn}=1$, and its singular values range from $1$ to $1/\rho$ for $\rho\in\{1,4,16\}$. The second penalty therefore equals the separated product. Each method independently selects its regularization parameter by validation. An orthonormalized implementation of the composed estimator provides an equivalence check. We use $16$ paired seeds, $6$ tasks, $40$ training and $80$ validation observations per task, and true coefficient matrices of rank two with equal population signal strength. The signal lies in directions associated with the largest singular values of $K$, the smallest, or mixtures of the two. We evaluate excess population risk exactly as $\|\widehat F-F^\star\|_{\HS}^2/6$.

\begin{figure}[!t]
\centering
\includegraphics[width=\linewidth]{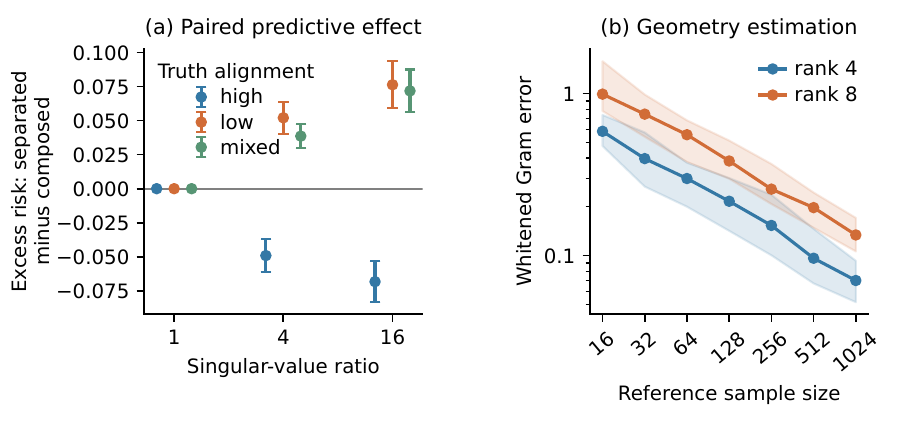}
\caption{Regularization and Gram estimation with fixed nonlinear features. (a) Excess error for the separated penalty minus excess error for the composed penalty, with descriptive $95\%$ Student-$t$ intervals over $16$ paired seeds in each cell. Positive values favor the composed penalty. (b) Median operator norm of the error in the whitened $L^2$ feature Gram matrix over $32$ independent repetitions; shading shows the $10$th--$90$th percentiles. Within each repetition, larger reference samples contain the smaller samples. Orthonormalizing the features gives an equivalent implementation of the composed penalty.}
\label{fig:composition-pilot}
\end{figure}

Figure~\ref{fig:composition-pilot}(a) shows that alignment determines which penalty benefits prediction. At $\rho=16$, the composed penalty reduces mean excess by $0.07631$ when the signal lies in the least amplified directions and $0.07180$ for the mixed signal; the corresponding descriptive $95\%$ paired-seed intervals are $[0.05902,0.09360]$ and $[0.05607,0.08752]$. When the signal lies in the most amplified directions, the separated penalty reduces excess by $0.06829$ $[0.05326,0.08333]$. It preferentially suppresses directions that contain only noise in this case. For each fixed signal, the composed estimator gives identical fitted functions as $\rho$ varies, and its orthonormalized implementation agrees exactly. The isotropic comparison also agrees exactly. All $7{,}344$ fitted candidates satisfy the declared relative primal--dual gap tolerance $10^{-7}$.

\subsection{Accuracy of the estimated Gram matrix}\label{sec:geometry-accuracy}

\paragraph{Independent reference inputs.}
A separate study uses independent reference inputs to estimate feature Gram matrices of ranks $4$ and $8$, including singular $Q$. After whitening, the leverage bound equals the rank. Over $32$ independent repetitions, the median spectral error at rank $8$ decreases from $0.992$ with $16$ reference inputs to $0.134$ with $1{,}024$; at rank $4$, it decreases from $0.585$ to $0.070$. Figure~\ref{fig:composition-pilot}(b) shows all reference sample sizes in the protocol. Panel (a) addresses the predictive comparison in \Cref{sec:composition-pilot}; panel (b) addresses the accuracy of the estimated geometry.

\Cref{app:composition-pilot} gives the complete design, all cell means,
the sampling of reference inputs and the numerical certificates.

\paragraph{SARCOS application.}
\label{sec:sarcos-application}

We compare exact and estimated geometry on SARCOS, using 21 measurements
of robot state to model seven joint torques \citep{SARCOSGPMLData}.
The three methods select among the same three 16-feature subspaces of a
fixed 48-function sinusoidal dictionary, with equal validation budgets.
Its inner product uses a standard Gaussian reference law in independently
standardized input coordinates. This law need not match the robot's input
distribution. We compare the exact Gram, an estimate from 2,048 independent
reference points, and the Euclidean penalty on task coefficients, with
64 or 256 fitting observations per task and five paired training seeds.

Evaluation uses the 4,449 rows in the official test file. All inputs and
4,446 complete observations also occur in the official training file.
In individual runs, 332--346 evaluation observations at $n=64$ and
465--473 at $n=256$ also enter preparation or a task's training or validation
set. These scores therefore measure agreement on overlapping observations.
\Cref{app:sarcos-application} documents the overlap and selection procedure.

\Cref{tab:sarcos-application} reports estimated minus exact mean error as
the primary paired comparison. The errors closely match, and all methods
select the same subspace in each seed; this does not imply identical fitted
predictions. As a secondary comparison, separated minus exact error is
$0.00488\;[-0.00252,0.01227]$ at $n=64$ and
$0.00130\;[-0.00088,0.00348]$ at $n=256$, with descriptive $95\%$
paired intervals that include zero.

The observed approximation uses 2,048 reference inputs. The sufficient
reference-sample bound in \Cref{eq:feature-reference-size} is 189,521 for
$\varepsilon=1/2$, $\delta=0.05$ and the leverage bound used here.
Thus this application measures realized approximation accuracy without
certifying that prescribed confidence level. \Cref{app:sarcos-application}
gives the leverage calculation, observed spectral bounds, full protocol
and certificates for all 1,530 fits.

\begin{table}[H]
\centering
\small
\caption{SARCOS: mean standardized MSE over all 4,449 evaluation rows, including
observations used in preparation, training or validation, averaged across
seven torques and five paired seeds. The final column gives estimated minus
exact mean error and its descriptive 95\% paired interval across training
seeds, using the recorded paired differences rather than the rounded means.}
\label{tab:sarcos-application}
\begin{tabular}{rcccc}
\toprule
$n$ & Exact Gram & Estimated Gram & Separated & \shortstack{Estimated minus exact\\mean [95\% interval]} \\
\midrule
64 & 0.65904 & 0.65873 & 0.66392 & $-0.000309\;[-0.000709,0.000092]$ \\
256 & 0.57404 & 0.57407 & 0.57534 & $0.000033\;[-0.000242,0.000307]$ \\
\bottomrule
\end{tabular}
\end{table}

\FloatBarrier
\subsection{Unequal task counts}\label{sec:legacy-summary}
The controlled linear study comprises $252$ paired scenarios across
two suites. It fits the composed predictor directly and compares weights based on actual, equal or cyclically shifted task counts, alongside weighted Frobenius and ridge penalties. Correct
weighting reduces average population excess over unweighted nuclear
regularization by $0.00764$, with grouped-bootstrap $95\%$ interval
$[0.00465,0.01110]$. Against shifted counts, it improves both task-average
and least-sampled-quartile excess in all $15$ imbalanced cells. The
least-sampled-quartile comparison with unweighted nuclear regularization
varies across suites. \Cref{app:legacy-experiments} gives the design, results and
checks of the archived results.

\FloatBarrier

\section{Discussion and conclusion}\label{sec:conclusion}

The weighted trace norm characterizes the best separated bound over coordinates in a finite-dimensional intermediate space. The task Gram and standard norm duality connect this characterization to regularization and statistical complexity.

Under bounded leverage, uniform Gram approximation in a fixed feature span yields fitting guarantees and generalization under an estimated norm constraint.

Signal alignment can favor either penalty. Estimated and exact Grams give similar evaluation errors on overlapping SARCOS observations.

Open questions concern nonlinear features learned beyond a fixed span and coordinate changes realizable by a specified neural architecture.

\FloatBarrier

\Needspace{7\baselineskip}
\section*{Data and code availability}
The computational supplement contains protocols, code, configurations, synthetic data, fitted coefficient paths, validation selections, reference samples, numerical certificates, and recovered task-count outputs and protocol records. For SARCOS, the supplement gives official download links, checksums and split indices; \Cref{app:sarcos-application} documents the overlap between training and test files. \Cref{app:composition-pilot,app:release-status} describe the composition study and archived results.

\section*{AI Disclosure}
ChatGPT assisted with language editing, organization, mathematical development, proof checks, literature searches, and experimental design and code. The author remains responsible for the manuscript's claims, references, code and results.

\bibliography{main}

\Needspace{8\baselineskip}
\appendix
\pdfbookmark[0]{Appendices}{appendices}
\renewcommand{\theHfigure}{app.\arabic{figure}}
\renewcommand{\theHtable}{app.\arabic{table}}

\section{Core proofs}\label{app:proofs}

The proofs follow the main argument: intermediate coordinates, empirical complexity, Gram approximation and fitting, then population risk.

\noindent\textbf{Appendix roadmap.}
\Cref{sec:joint-covariance,sec:no-go,sec:pullback} give the covariance and
operator constructions, separation mechanisms, and realizations using
function values and derivatives. The experimental appendices follow the
main comparisons: composition and Gram estimation
(\Cref{app:composition-pilot}), SARCOS (\Cref{app:sarcos-application}),
and task-count regression (\Cref{app:legacy-experiments}).

\subsection{Coordinate characterization}\label{app:proof-coordinate}

\Needspace{4\baselineskip}
Proof of \Cref{thm:optimal-refactor}.

For the optimal intermediate-refactorization result, use the polar decomposition
$K=UA$, where $A=\left(K^*K\right)^{1/2}$ and $U$ is a partial isometry.  Let $P_0$ be the orthogonal projection onto $\ker K$ and define
\begin{align*}
R_\epsilon
=
A+\epsilon P_0,
\qquad
\epsilon>0.
\end{align*}
In finite dimension, $R_\epsilon$ is invertible and
$KR_\epsilon^{-1}=UP_{(\ker K)^\perp}$, whose operator norm is at most one.  With $C=G\Dn^{-1/2}$,
\begin{align*}
\inf_{R\in\operatorname{GL}(\mathcal H_L)}
\left\lVert KR^{-1}\right\rVert_{\opn}
\left\lVert RC\right\rVert_*
&\le
\lim_{\epsilon\downarrow0}
\left\lVert
R_\epsilon C
\right\rVert_*
\notag\\
&=
\left\lVert AC\right\rVert_*
=
\left\lVert UAC\right\rVert_*
=
\left\lVert KC\right\rVert_*.
\end{align*}
The penultimate equality holds because $AC$ lies in the initial space of $U$, where the partial isometry preserves singular values.  The reverse inequality follows from the ideal property for every invertible $R$.  If $K$ is injective, $A$ is invertible and $R=A$ attains the infimum.  This proves \Cref{thm:optimal-refactor}.
\hfill $\blacksquare$

\subsection{Exact empirical complexity and the Hilbert--Schmidt gap}\label{app:proof-complexity}

\begin{corollary}[Gap in the Hilbert--Schmidt relaxation]\label{cor:orthogonal-sharpness}
Assume that the representers
$\{k_0(\cdot,x_{ti})\}_{t,i}$ are pairwise orthogonal in $\mathcal H_0$ and satisfy
$k_0(x_{ti},x_{ti})=\kappa$.  Then
\begin{align}
\taskR\left(\mBjoint(\tau)\right)
=
\frac{\tau\sqrt{\kappa}}{\Task}.
\label{eq:orthogonal-exact}
\end{align}
Consequently, the general Hilbert--Schmidt upper bound in \Cref{eq:hs-relaxation} is larger by the factor $\sqrt{\Task}$ in this construction.
\end{corollary}

\Needspace{4\baselineskip}
Proof of \Cref{thm:exact-rademacher,cor:orthogonal-sharpness}.

For a fixed realization of the Rademacher variables, the reproducing property and \Cref{eq:sample-operator} give
\begin{align}
\frac{1}{\Task}
\sum_{t=1}^{\Task}
\frac{1}{n_t}
\sum_{i=1}^{n_t}
\varepsilon_{ti}
\left(Fe_t\right)\left(x_{ti}\right)
&=
\frac{1}{\Task}
\sum_{t=1}^{\Task}
\left\langle
z_t,Fe_t
\right\rangle_{\mathcal H_0}
\notag\\
&=
\frac{1}{\Task}
\operatorname{Tr}\left(Z_\varepsilon^*F\right).
\label{eq:proof-trace-pairing}
\end{align}
Set $C=F\Dn^{-1/2}$, so $F=C\Dn^{1/2}$.  Cyclicity of the finite-dimensional trace yields
\begin{align}
\operatorname{Tr}\left(Z_\varepsilon^*F\right)
&=
\operatorname{Tr}
\left(
\Dn^{1/2}Z_\varepsilon^*C
\right)
\notag\\
&=
\left\langle
Z_\varepsilon\Dn^{1/2},C
\right\rangle_{\HS}.
\label{eq:proof-dual-pairing}
\end{align}
Because the dual norm of the nuclear norm is the operator norm,
\begin{align}
\sup_{\lVert C\rVert_*\le\tau}
\left\langle
Z_\varepsilon\Dn^{1/2},C
\right\rangle_{\HS}
=
\tau
\left\lVert
Z_\varepsilon\Dn^{1/2}
\right\rVert_{\opn}.
\label{eq:proof-nuclear-duality}
\end{align}
The equality is attained by a rank-one nuclear-norm extremizer aligned with top left and right singular vectors of the sample operator.  Combining \Cref{eq:proof-trace-pairing,eq:proof-dual-pairing,eq:proof-nuclear-duality}, dividing by $\Task$, and taking the Rademacher expectation proves \Cref{eq:exact-rademacher}.

For the simpler upper bound,
\begin{align}
\mathbb E_\varepsilon
\left\lVert
Z_\varepsilon\Dn^{1/2}
\right\rVert_{\opn}
&\le
\left[
\mathbb E_\varepsilon
\left\lVert
Z_\varepsilon\Dn^{1/2}
\right\rVert_{\HS}^2
\right]^{1/2}
\notag\\
&=
\left[
\sum_{t=1}^{\Task}
n_t\mathbb E_\varepsilon
\left\lVert z_t\right\rVert_{\mathcal H_0}^2
\right]^{1/2}.
\label{eq:proof-hs-step}
\end{align}
Independence and centering of the signs imply
\begin{align*}
\mathbb E_\varepsilon
\left\lVert z_t\right\rVert_{\mathcal H_0}^2
=
\frac{1}{n_t^2}
\sum_{i=1}^{n_t}
k_0\left(x_{ti},x_{ti}\right)
\le
\frac{\kappa}{n_t}.
\end{align*}
Substitution into \Cref{eq:proof-hs-step} gives
$\mathbb E\lVert Z_\varepsilon\Dn^{1/2}\rVert_{\opn}\le\sqrt{\kappa\Task}$ and proves \Cref{eq:hs-relaxation}.

Under the assumptions of \Cref{cor:orthogonal-sharpness},
$\sqrt{n_t}z_t$ has norm $\sqrt\kappa$ for every sign realization, and the columns
$\{\sqrt{n_t}z_t\}_{t=1}^{\Task}$ are mutually orthogonal.  Therefore
$\lVert Z_\varepsilon\Dn^{1/2}\rVert_{\opn}=\sqrt\kappa$ deterministically.  Substitution into \Cref{eq:exact-rademacher} proves \Cref{eq:orthogonal-exact}.
\hfill $\blacksquare$

\subsection{Separated constraints}\label{app:proof-separated}

\begin{corollary}[Separated constraints as a relaxation]\label{cor:separated-relaxation}
Let $\mathcal A$ be any set of pairs $(\theta,\bg)$ such that
\begin{align*}
\lVert\KT\rVert_{\opn}
\le
\Gamma,
\qquad
\left\lVert
\Gg\Dn^{-1/2}
\right\rVert_*
\le
\rho.
\end{align*}
Then the corresponding end-to-end maps satisfy
\begin{align}
\mJ\left(\Ftg\right)
\le
\Gamma\rho,
\label{eq:ideal-relaxation}
\end{align}
and their empirical Rademacher complexity is at most
\begin{align}
\frac{\Gamma\rho}{\Task}
\mathbb E_\varepsilon
\left\lVert
Z_\varepsilon\Dn^{1/2}
\right\rVert_{\opn}
\le
\Gamma\rho
\sqrt{\frac{\kappa}{\Task}}.
\label{eq:separated-corollary}
\end{align}
\end{corollary}

\Needspace{4\baselineskip}
Proof of \Cref{cor:separated-relaxation}.

The ideal property of the nuclear norm gives
\begin{align*}
\left\lVert
\KT\Gg\Dn^{-1/2}
\right\rVert_*
\le
\left\lVert\KT\right\rVert_{\opn}
\left\lVert
\Gg\Dn^{-1/2}
\right\rVert_*
\le
\Gamma\rho.
\end{align*}
The realizable class is therefore contained in $\mBjoint(\Gamma\rho)$.  Apply \Cref{thm:exact-rademacher}.
\hfill $\blacksquare$

\subsection{Uniform Gram approximation and fitting}\label{app:estimated-feature-geometry}

We first prove \Cref{thm:estimated-feature-geometry} and the containment in \Cref{eq:estimated-ball-containment}, then give the complete optimization consequences.

\paragraph{Nullspaces and restriction to the represented range.}
Let $v_1,\ldots,v_{p-r}$ be an orthonormal basis of $\ker Q$.
For each $v_j$,
\begin{align*}
\mathbb E\bigl[v_j^\top W(U)v_j\bigr]=v_j^\top Qv_j=0.
\end{align*}
The random variable inside the expectation is nonnegative.  It is
therefore zero almost surely.  Since $W(U)$ is positive semidefinite,
$v_j^\top W(U)v_j=\lVert W(U)^{1/2}v_j\rVert_2^2=0$ implies
$W(U)v_j=0$.  There are only finitely many basis vectors, so one
probability-one event gives $\ker Q\subseteq\ker W(U)$.
Consequently $W(U)=P W(U)P$ almost surely, where $P$ is the orthogonal
projector onto $\operatorname{range}(Q)$; this conclusion uses symmetry
to eliminate the off-diagonal blocks.  It also holds for every summand
and for $\widehat Q$ on a common probability-one event.
If $r=0$, the same argument proves the asserted degenerate case.

Assume henceforth that $r\ge1$.  Choose a $p\times r$ matrix $V$ with
orthonormal columns spanning $\operatorname{range}(Q)$ and put
$Q_r=V^\top QV\succ0$.  Define
\begin{align*}
X_j=Q_r^{-1/2}V^\top W(U_j)VQ_r^{-1/2}
\qquad(j=1,\ldots,m).
\end{align*}
These are independent positive-semidefinite $r\times r$ random matrices,
and
\begin{align*}
\mathbb EX_j=I_r,
\qquad 0\preceq X_j\preceq L I_r.
\end{align*}
The upper bound follows by congruence from
\Cref{eq:feature-leverage-bound}.  Conversely, a bound by $LI_r$ for
the whitened restriction, together with the nullspace property just
proved, implies $W(U)\preceq LQ$.  Since
$Q^{\dagger/2}=VQ_r^{-1/2}V^\top$, this also proves the stated
equivalence with the pseudoinverse expression.  Taking expectations in
$X_j\preceq LI_r$ shows that necessarily $L\ge1$ when $r>0$.

\paragraph{Matrix concentration and the constant.}
Apply \citet[Corollary~5.2]{Tropp2012UserFriendly} to $X_1,\ldots,X_m$.
The matrix dimension is $r$, the summand bound is $L$, and both extreme
eigenvalues of $\sum_j\mathbb EX_j$ equal $m$.  With
$\overline X=m^{-1}\sum_jX_j$ and $0<\varepsilon<1$, the resulting
two bounds are
\begin{align*}
\mathbb P\{\lambda_{\min}(\overline X)\le1-\varepsilon\}
&\le r\exp\!\left[-\frac mL
 \{\varepsilon+(1-\varepsilon)\log(1-\varepsilon)\}\right],
\\
\mathbb P\{\lambda_{\max}(\overline X)\ge1+\varepsilon\}
&\le r\exp\!\left[-\frac mL
 \{(1+\varepsilon)\log(1+\varepsilon)-\varepsilon\}\right].
\end{align*}
For $0\le t<1$, $-\log(1-t)\ge t$.  Integrating from $0$ to
$\varepsilon$ gives
$\varepsilon+(1-\varepsilon)\log(1-\varepsilon)\ge\varepsilon^2/2$.
For $0\le t\le1$, the inequality
$\log(1+t)\ge t-t^2/2$ follows because the difference has derivative
$t^2/(1+t)\ge0$ and is zero at $0$.  Integrating gives
\begin{align*}
(1+\varepsilon)\log(1+\varepsilon)-\varepsilon
\ge\frac{\varepsilon^2}{2}-\frac{\varepsilon^3}{6}
\ge\frac{\varepsilon^2}{3}.
\end{align*}
The union bound therefore shows that
\begin{align*}
\mathbb P\{(1-\varepsilon)I_r\preceq\overline X
                    \preceq(1+\varepsilon)I_r\}
\ge1-2r\exp\!\left(-\frac{m\varepsilon^2}{3L}\right).
\end{align*}
The event is measurable because it is specified by the extreme
eigenvalues of a finite random matrix.  Under
\Cref{eq:feature-reference-size}, its probability is at least $1-\delta$.
Congruence by $VQ_r^{1/2}$ gives
\Cref{eq:feature-spectral-sandwich}: the middle matrix is $\widehat Q$
by the common nullspace property.  The two inequalities, with
$1-\varepsilon>0$, also imply $\ker\widehat Q=\ker Q$.

\paragraph{From the feature Gram to every task head.}
Work on the preceding event and fix any
$A\in\mathbb R^{p\times\Task}$.  Set $B=A\Dn^{-1/2}$.
Congruence by $B$ gives
\begin{align*}
(1-\varepsilon)B^\top QB
\preceq B^\top\widehat Q B
\preceq(1+\varepsilon)B^\top QB.
\end{align*}
For positive-semidefinite matrices $M\preceq N$, the min--max
characterization of ordered eigenvalues gives
$\lambda_j(M)\le\lambda_j(N)$ for every $j$.  Indeed,
$x^\top Mx\le x^\top Nx$ for every unit vector $x$, so the inequality
is preserved by the maxima and minima over the subspaces in that
characterization.  Applying the increasing scalar square root and
summing the eigenvalues yields
$\operatorname{Tr}(M^{1/2})\le\operatorname{Tr}(N^{1/2})$.
Applying this fact to the two Gram inequalities proves
\Cref{eq:feature-penalty-sandwich}.  The event did not depend on $A$,
so the conclusion holds simultaneously for every head matrix, without
a union bound or a measurability assumption on a supremum over heads.
The two set inclusions in \Cref{eq:estimated-ball-containment} follow
by applying the upper and lower penalty bounds, respectively.
\hfill $\blacksquare$

\subsubsection{Regularized objective comparison}\label{app:objective-comparison}

\begin{corollary}[Regularized objective comparison]
\label{cor:estimated-geometry-objective}
On the event in \Cref{eq:feature-spectral-sandwich}, write
$c_-=\sqrt{1-\varepsilon}$ and $c_+=\sqrt{1+\varepsilon}$.
Let $\mathcal A\subseteq\mathbb R^{p\times\Task}$ be nonempty and let
$\widehat L:\mathcal A\to[0,+\infty]$ be finite at some member of
$\mathcal A$.  Fix $\lambda\ge0$ and $\eta\ge0$, and suppose
$\widehat A\in\mathcal A$ satisfies
\begin{align}
\widehat L(\widehat A)+\lambda\widehat J(\widehat A)
\le\inf_{A\in\mathcal A}
\{\widehat L(A)+\lambda\widehat J(A)\}+\eta.
\label{eq:approximate-estimated-minimizer}
\end{align}
Then, for every $A\in\mathcal A$,
\begin{align}
\widehat L(\widehat A)+\lambda c_-J_Q(\widehat A)
\le\widehat L(A)+\lambda c_+J_Q(A)+\eta.
\label{eq:estimated-objective-comparator}
\end{align}
In particular, for
$\mathcal E_\lambda(A)=\widehat L(A)+\lambda J_Q(A)$,
\begin{align}
\mathcal E_\lambda(\widehat A)
\le\frac{c_+}{c_-}\inf_{A\in\mathcal A}\mathcal E_\lambda(A)
+\frac{\eta}{c_-}.
\label{eq:estimated-objective-factor}
\end{align}
No convexity or attainment of the infimum is required.  For $\eta=0$,
\Cref{eq:approximate-estimated-minimizer} assumes an actual minimizer exists.
\end{corollary}

\paragraph{Proof of \Cref{cor:estimated-geometry-objective}.}\label{app:proof-objective}

For every $A\in\mathcal A$, the approximate minimizer assumption and
the penalty sandwich imply
\begin{align*}
\widehat L(\widehat A)+\lambda c_-J_Q(\widehat A)
&\le\widehat L(\widehat A)+\lambda\widehat J(\widehat A)
\\
&\le\widehat L(A)+\lambda\widehat J(A)+\eta
\\
&\le\widehat L(A)+\lambda c_+J_Q(A)+\eta.
\end{align*}
This proves \Cref{eq:estimated-objective-comparator}.  Because
$0<c_-\le1\le c_+$ and the loss is nonnegative, the left side is at
least $c_-\mathcal E_\lambda(\widehat A)$, while the right side is
at most $c_+\mathcal E_\lambda(A)+\eta$.  Divide by $c_-$ and take
the infimum over $A$ to obtain \Cref{eq:estimated-objective-factor}.
Taking this infimum uses no minimizing element.  Since the objectives
are nonnegative and finite at some $A$, their infima are finite, and
an $\eta$-approximate minimizer exists for every $\eta>0$ by the
definition of infimum.  No existence is asserted for $\eta=0$.
All inequalities hold on the same measurable Gram event for every
head matrix.  They therefore also hold pointwise for any selected
heads and any chosen head set satisfying the displayed optimization
condition.  Conditional application after independently learning the
features follows by applying the argument with those features fixed.
\hfill $\blacksquare$\par
This comparison concerns the same empirical loss evaluated with an
estimated or exact geometry.  The admissible task coefficients may satisfy architectural or
rank constraints.  The feature family may first be learned from separate
data: conditioning on that data gives the same result whenever the stated
assumptions hold conditionally.  Selecting features outside the fixed
span using the reference sample requires a uniform event over those
candidate families.  Within the fixed span, the same Gram event also
controls learned representations, as follows.

\subsubsection{Shared linear maps in a fixed span}\label{app:adaptive-feature-span}

\begin{corollary}[Learning representations within a fixed feature span]
\label{cor:adaptive-feature-span}
On the event in \Cref{eq:feature-spectral-sandwich}, simultaneously
for every integer $q\ge1$ and every $M\in\mathbb R^{p\times q}$,
\begin{align}
(1-\varepsilon)M^\top QM
\preceq M^\top\widehat Q M
\preceq(1+\varepsilon)M^\top QM.
\label{eq:adaptive-representation-gram}
\end{align}
For every $A\in\mathbb R^{q\times\Task}$, the exact and estimated
penalties of the composed predictor $F_{M,A}=KMA$ are
$J_Q(MA)$ and $J_{\widehat Q}(MA)$, and satisfy
\Cref{eq:feature-penalty-sandwich} with $MA$ in place of $A$.

Let $\mathcal C$ be any nonempty collection of such pairs $(M,A)$,
allowing $q$ to vary, and let
$\widehat L:\mathcal C\to[0,+\infty]$ be finite at some pair.
Fix $\lambda\ge0$ and $\eta\ge0$.  If
$(\widehat M,\widehat A)\in\mathcal C$ satisfies
\begin{align*}
&\widehat L(\widehat M,\widehat A)
 +\lambda J_{\widehat Q}(\widehat M\widehat A)
\\
&\quad\le\inf_{(M,A)\in\mathcal C}
 \{\widehat L(M,A)+\lambda J_{\widehat Q}(MA)\}+\eta,
\end{align*}
then, on the same event,
\begin{align}
&\widehat L(\widehat M,\widehat A)
 +\lambda c_-J_Q(\widehat M\widehat A)
\notag\\
&\quad\le\inf_{(M,A)\in\mathcal C}
 \{\widehat L(M,A)+\lambda c_+J_Q(MA)\}+\eta,
\label{eq:adaptive-span-objective}
\end{align}
where $c_-=\sqrt{1-\varepsilon}$ and $c_+=\sqrt{1+\varepsilon}$.
The same-parameter objective bound
\Cref{eq:estimated-objective-factor} holds with the infimum over
$\mathcal C$.  Both fitted matrices may depend on the reference sample.
\end{corollary}

\paragraph{Proof of \Cref{cor:adaptive-feature-span}.}\label{app:proof-adaptive-span}

Congruence of \Cref{eq:feature-spectral-sandwich} by any
$M\in\mathbb R^{p\times q}$ proves
\Cref{eq:adaptive-representation-gram}.  Moreover, the two matrices
$Q^{1/2}MA\Dn^{-1/2}$ and
$(M^\top QM)^{1/2}A\Dn^{-1/2}$ have the same task Gram,
namely $\Dn^{-1/2}A^\top M^\top QMA\Dn^{-1/2}$.
Their nuclear norms are therefore equal, including when $M$ or $Q$
is singular.  The same reasoning applies to $\widehat Q$.
The penalty sandwich follows by taking $B=MA$ in the already uniform
head bound.  This is a deterministic consequence of the original
Gram event for every width, representation matrix and head matrix;
no union bound over these choices is used.

For the chosen $\lambda$, approximate minimization and the uniform
penalty inequalities give, for each $(M,A)\in\mathcal C$,
\begin{align*}
&\widehat L(\widehat M,\widehat A)
 +\lambda c_-J_Q(\widehat M\widehat A)
\\
&\quad\le\widehat L(\widehat M,\widehat A)
 +\lambda J_{\widehat Q}(\widehat M\widehat A)
\\
&\quad\le\widehat L(M,A)+\lambda J_{\widehat Q}(MA)+\eta
\\
&\quad\le\widehat L(M,A)+\lambda c_+J_Q(MA)+\eta.
\end{align*}
Taking the infimum gives \Cref{eq:adaptive-span-objective}.
The same nonnegative-loss comparison used in
\Cref{cor:estimated-geometry-objective} gives the same-parameter
objective factor over $\mathcal C$.  Neither the argument nor
approximate minimization requires a compact or convex parameter set.
\hfill $\blacksquare$\par
The matrix $M$ selects or combines the fixed dictionary functions;
its width, rank and coefficients may be chosen during fitting.
The event covers all such choices because their composed predictors
remain in the same feature span.  Learning nonlinear features outside
that span requires an additional uniform approximation argument.

\subsubsection{Several prespecified dictionaries}\label{app:multiple-dictionaries}

For $N$ separately prespecified dictionaries satisfying the theorem's
assumptions with ranks $r_\ell$ and bounds $L_\ell$, a union bound gives
simultaneous control whenever
$m\ge3L_{\max}\varepsilon^{-2}\log(2\sum_\ell r_\ell/\delta)$,
where $L_{\max}=\max_\ell L_\ell$, provided the rank sum is positive;
if all ranks are zero, every Gram is exact almost surely.
The proof does not require independence between dictionaries.

\paragraph{Proof of the dictionary extension.}\label{app:proof-multiple-dictionaries}

For the separate-dictionary statement, the nonzero-rank dictionary
$\ell$ has failure probability at most
$2r_\ell\exp(-m\varepsilon^2/(3L_\ell))$ by the theorem's proof;
a zero-rank member is exact almost surely.  The sum of failure
probabilities is at most
\begin{align*}
2\Bigl(\sum_{\ell=1}^N r_\ell\Bigr)
 \exp\!\left(-\frac{m\varepsilon^2}{3L_{\max}}\right),
\end{align*}
which proves the stated threshold.  A finite union bound applies even
when dictionaries use the same reference points.  It also gives
simultaneous control for subsequent selection between the dictionaries.
When they represent predictors, their Grams must correspond to the
same prescribed function inner product to compare penalties on a
common scale.
\hfill $\blacksquare$

\subsection{Population-risk consequences}\label{app:population}

For a predictor map $F$, define the task-balanced population and empirical risks
\begin{align*}
\popR(F)
&=
\frac{1}{\Task}
\sum_{t=1}^{\Task}
\mathbb E_{(X,Y)\sim P_t}
\loss_t\left(\left(Fe_t\right)(X),Y\right),
\\
\empR(F)
&=
\frac{1}{\Task}
\sum_{t=1}^{\Task}
\frac{1}{n_t}
\sum_{i=1}^{n_t}
\loss_t\left(\left(Fe_t\right)(x_{ti}),y_{ti}\right).
\end{align*}

\begin{theorem}[Population bound for the joint class]\label{thm:population}
Under the sampling and measurable RKHS setting of \Cref{sec:setting}, fix $\tau\ge0$ before sampling and assume each loss is jointly measurable, $L_\loss$-Lipschitz in its prediction argument, and takes values in $[0,B_\loss]$.  For every $\delta\in(0,1)$, with probability at least $1-\delta$, every $F\in\mBjoint(\tau)$ satisfies
\begin{align}
\popR(F)
&\le
\empR(F)
+
\frac{2L_\loss\tau}{\Task}
\mathbb E_\varepsilon
\left\lVert
Z_\varepsilon\Dn^{1/2}
\right\rVert_{\opn}
\notag\\*
&\quad+
3B_\loss
\left[
\frac{\log(2/\delta)}{2\Task^2}
\sum_{t=1}^{\Task}
\frac{1}{n_t}
\right]^{1/2}.
\label{eq:population-bound}
\end{align}
Under \Cref{eq:kernel-bound}, the middle term is at most
$2L_\loss\tau\sqrt{\kappa/\Task}$.
\end{theorem}

\paragraph{Proof of \Cref{thm:population}.}\label{app:proof-population}

Separability gives a countable dense subset of $\mBjoint(\tau)$ in the finite-product Hilbert norm. Bounded evaluation and the Lipschitz loss make empirical evaluations and population risk continuous on this class, so its suprema equal suprema over that subset and are measurable. Let $w_{ti}=1/(\Task n_t)$ and let $\loss\circ\mBjoint(\tau)$ be the induced loss class.  Ghost-sample symmetrization gives
\begin{align*}
\mathbb E_S
\sup_{F\in\mBjoint(\tau)}
\left\{
\popR(F)-\empR(F)
\right\}
\le
2\mathbb E_S
\widehat{\mathfrak R}^{\loss}_{\bn}
\left(\loss\circ\mBjoint(\tau)\right).
\end{align*}
Center each scalar loss at zero and apply the contraction inequality:
\begin{align*}
\widehat{\mathfrak R}^{\loss}_{\bn}
\left(\loss\circ\mBjoint(\tau)\right)
\le
L_\loss
\taskR\left(\mBjoint(\tau)\right).
\end{align*}
Replacing observation $(t,i)$ changes the population-gap supremum and the empirical loss-class complexity by at most $B_\loss w_{ti}$.  Two applications of McDiarmid's inequality, each with failure probability $\delta/2$, followed by a union bound yield, simultaneously for every $F$,
\begin{align*}
\popR(F)
\le
\empR(F)
+
2L_\loss
\taskR\left(\mBjoint(\tau)\right)
+
3B_\loss
\left[
\frac{\log(2/\delta)}{2}
\sum_{t,i}w_{ti}^2
\right]^{1/2}.
\end{align*}
Since
$\sum_{t,i}w_{ti}^2=\Task^{-2}\sum_t n_t^{-1}$, substituting \Cref{eq:exact-rademacher} proves \Cref{eq:population-bound}.
\hfill $\blacksquare$

\subsubsection{Using an estimated norm constraint}

\begin{corollary}[Population control from an estimated norm constraint]
\label{cor:estimated-radius-population}
Assume the hypotheses of \Cref{thm:estimated-feature-geometry} and
\Cref{thm:population}.  In particular, the input RKHS, task counts and
sampling distributions are fixed before the labelled sample is drawn,
and the losses are measurable, $L_\loss$-Lipschitz and bounded in
$[0,B_\loss]$.  The dictionary is fixed independently of the reference
sample, which is independent of the labelled sample.  Fix
$\tau\ge0$, $0<\varepsilon<1$, and
$\delta_{\mathrm g},\delta_{\mathrm r}\in(0,1)$ with
$\delta_{\mathrm g}+\delta_{\mathrm r}<1$ before sampling.
Choose $m$ by \Cref{eq:feature-reference-size} with
$\delta=\delta_{\mathrm g}$, using its zero-rank alternative when
appropriate.  Then, with probability at least
$1-\delta_{\mathrm g}-\delta_{\mathrm r}$, every fitted pair
$(M,A)$ in the fixed dictionary span satisfying
$J_{\widehat Q}(MA)\le\tau$ has composed map $F=KMA$ satisfying
\begin{align}
\popR(F)
&\le\empR(F)
 +\frac{2L_\loss\tau}{\Task\sqrt{1-\varepsilon}}
   \mathbb E_\varepsilon
   \lVert Z_\varepsilon\Dn^{1/2}\rVert_{\opn}
\notag\\
&\quad+3B_\loss
 \left[\frac{\log(2/\delta_{\mathrm r})}{2\Task^2}
              \sum_{t=1}^{\Task}\frac1{n_t}\right]^{1/2}.
\label{eq:estimated-radius-population}
\end{align}
Here the expectation in the middle term is over the Rademacher signs
used in \Cref{thm:population}.  The selected representation matrix and
task heads may depend on both samples.
\end{corollary}

\begin{proof}\label{app:proof-estimated-radius}
On the Gram event, the penalty sandwich gives
$\mJ(F)=J_Q(MA)\le\tau/\sqrt{1-\varepsilon}$ for every admissible
pair.  This radius is deterministic.  Apply \Cref{thm:population} at
that radius and confidence level $\delta_{\mathrm r}$.
The resulting event controls every map in the fixed ambient ball,
including all the selected pairs on the Gram event.  A union bound
over the two failure events proves the claim.  Uniformity over the
ambient ball is why selecting a representation within the fixed span
does not require an additional supremum in this argument.
\end{proof}

\FloatBarrier

\section{Task covariance, rank and operator geometry}\label{sec:joint-covariance}

\subsection{Task covariance and rank}\label{app:covariance-profile}

Applying the covariance formulation of output-kernel learning~\citep{dinuzzo2011lowrank,Ciliberto2015ConvexTasks} to the composed predictors gives the following profile. For a fixed predictor map $F$, set
\begin{align*}
S_F
=
F^*F
=
\left(
\left\langle
f_t,f_s
\right\rangle_{\mathcal H_0}
\right)_{t,s=1}^{\Task}.
\end{align*}
For $\Sigma\in\mathbb S_{++}^{\Task}$, define
\begin{align}
\mathfrak C_{\Sigma}^{\mathrm{end}}\left(F;\bn\right)
=
\left[
\operatorname{Tr}\left(\Dn^{-1}\Sigma\right)
\operatorname{Tr}\left(\Sigma^{-1}S_F\right)
\right]^{1/2}.
\label{eq:end-covariance-profile}
\end{align}

\begin{theorem}[End-to-end covariance profile]\label{thm:end-covariance}
For every bounded $F:\mathbb R^\Task\to\mathcal H_0$,
\begin{align}
\inf_{\Sigma\in\mathbb S_{++}^{\Task}}
\mathfrak C_{\Sigma}^{\mathrm{end}}\left(F;\bn\right)
=
\operatorname{Tr}
\left[
\left(
\Dn^{-1/2}S_F\Dn^{-1/2}
\right)^{1/2}
\right]
=
\mJ(F).
\label{eq:end-covariance-identity}
\end{align}
If $\Dn^{-1/2}S_F\Dn^{-1/2}\succ0$, every minimizer is a positive scalar multiple of
\begin{align}
\Sigma_F^\star
=
\Dn^{1/2}
\left(
\Dn^{-1/2}S_F\Dn^{-1/2}
\right)^{1/2}
\Dn^{1/2}.
\label{eq:end-covariance-minimizer}
\end{align}
If the weighted Gram matrix is nonzero and singular, the value in \Cref{eq:end-covariance-identity} is an unattained infimum over the positive-definite cone, approached by the explicit regularized sequence in the proof.  If it is zero, the profile is zero for every positive-definite $\Sigma$.
\end{theorem}

\paragraph{Proof of \Cref{thm:end-covariance}.}\label{app:proof-covariance}

Set
\begin{align*}
B
=
\Dn^{-1/2}S_F\Dn^{-1/2}
\succeq
0,
\qquad
Q
=
\Dn^{-1/2}\Sigma\Dn^{-1/2}
\succ
0.
\end{align*}
Cyclicity of the trace gives
\begin{align*}
\operatorname{Tr}\left(\Dn^{-1}\Sigma\right)
=
\operatorname{Tr}(Q),
\qquad
\operatorname{Tr}\left(\Sigma^{-1}S_F\right)
=
\operatorname{Tr}\left(Q^{-1}B\right).
\end{align*}
Write
$B^{1/2}=Q^{1/2}\left(Q^{-1/2}B^{1/2}\right)$ and apply Hilbert--Schmidt Cauchy--Schwarz:
\begin{align}
\operatorname{Tr}\left(B^{1/2}\right)^2
\le
\operatorname{Tr}(Q)
\operatorname{Tr}\left(Q^{-1}B\right).
\label{eq:proof-covariance-lower}
\end{align}
If $B\succ0$, equality holds for $Q=cB^{1/2}$ with any $c>0$, which yields \Cref{eq:end-covariance-minimizer}.  If
$B=U\operatorname{diag}(\lambda_1,\ldots,\lambda_r,0,\ldots,0)U^\top$ with $0<r<\Task$, define
\begin{align*}
Q_\epsilon
=
U\operatorname{diag}
\left(
\sqrt{\lambda_1},\ldots,\sqrt{\lambda_r},
\epsilon,\ldots,\epsilon
\right)U^\top.
\end{align*}
Then $Q_\epsilon\succ0$,
$\operatorname{Tr}(Q_\epsilon^{-1}B)=\sum_{j=1}^r\sqrt{\lambda_j}$, and
$\operatorname{Tr}(Q_\epsilon)\to\sum_{j=1}^r\sqrt{\lambda_j}$.  Equality cannot be attained by a positive-definite $Q$ because equality in \Cref{eq:proof-covariance-lower} would require $Q$ to be proportional to singular $B^{1/2}$.  If $B=0$, the second trace is zero for every $Q\succ0$.

Finally,
\begin{align*}
\operatorname{Tr}\left(B^{1/2}\right)
=
\left\lVert
F\Dn^{-1/2}
\right\rVert_*,
\end{align*}
because $B=\left(F\Dn^{-1/2}\right)^*\left(F\Dn^{-1/2}\right)$.  This proves \Cref{thm:end-covariance}.
\hfill $\blacksquare$

For $F=\Ftg$, the Gram matrix in \Cref{eq:end-covariance-profile} is exactly \Cref{eq:end-to-end-gram}.  The optimized covariance therefore depends on the task functions after they have passed through the shared representation.  The next comparison recovers the separated product by bounding the Gram matrix of the composed functions using the Gram matrix of the terminal task functions.

\paragraph{The separated task profile is recovered as an upper bound.}
For any bounded $K:\mathcal H_L\to\mathcal H_0$, task map $G$, and $\Sigma\succ0$,
\begin{align}
\mathfrak C_{\Sigma}^{\mathrm{end}}(KG;\bn)
&\le
\lVert K\rVert_{\opn}
\left[
\operatorname{Tr}(\Dn^{-1}\Sigma)
\operatorname{Tr}(\Sigma^{-1}G^*G)
\right]^{1/2}.
\label{eq:separated-covariance-relaxation}
\end{align}
This follows from
$G^*K^*KG\preceq\lVert K\rVert_{\opn}^2G^*G$.
Optimizing \Cref{eq:separated-covariance-relaxation} over $\Sigma$ gives
\begin{align*}
\mJ(KG)
\le
\lVert K\rVert_{\opn}
\lVert G\Dn^{-1/2}\rVert_*,
\end{align*}
which is \Cref{eq:ideal-relaxation}.  This locates the loss from separation in the replacement of $K^*K$ by the isotropic upper bound $\lVert K\rVert_{\opn}^2I$.

\begin{corollary}[Rank bound for the end-to-end task map]\label{cor:joint-rank}
Let $r_F=\operatorname{rank}(F)$.  Then
\begin{align}
\mJ(F)
\le
\sqrt{r_F}
\left(
\sum_{t=1}^{\Task}
\frac{\lVert f_t\rVert_{\mathcal H_0}^2}{n_t}
\right)^{1/2}.
\label{eq:joint-rank-bound}
\end{align}
The rank in \Cref{eq:joint-rank-bound} is the dimension of the represented predictor family, which may be strictly smaller than the rank of the terminal task map $\Gg$.
\end{corollary}

\paragraph{Proof of \Cref{cor:joint-rank}.}\label{app:proof-rank}

For \Cref{cor:joint-rank}, set $C=F\Dn^{-1/2}$.  Cauchy--Schwarz for the nonzero singular values gives
\begin{align*}
\lVert C\rVert_*
\le
\sqrt{\operatorname{rank}(C)}
\lVert C\rVert_{\HS},
\qquad
\lVert C\rVert_{\HS}^2
=
\sum_{t=1}^{\Task}
\frac{\lVert f_t\rVert_{\mathcal H_0}^2}{n_t}.
\end{align*}
Since $\operatorname{rank}(C)=\operatorname{rank}(F)$, \Cref{eq:joint-rank-bound} follows.
\hfill $\blacksquare$

\subsection{Identifying terminal functions with the same predictor}\label{app:quotient-geometry}

A separate operator-level issue is whether distinct terminal functions induce the same predictor.  The following proposition identifies the equivalence classes on which the spectrum of the composed task map depends.

\begin{proposition}[Quotient-space factorization]\label{prop:quotient}
Let $K:\mathcal H_L\to\mathcal H_0$ be any bounded operator and let
$q:\mathcal H_L\to\mathcal H_L/\ker K$ be the quotient map.  There is a unique injective bounded operator
$\overline K:\mathcal H_L/\ker K\to\mathcal H_0$ such that
\begin{align}
K
=
\overline Kq,
\qquad
\left\lVert\overline K\right\rVert_{\opn}
=
\left\lVert K\right\rVert_{\opn}.
\label{eq:quotient-factorization}
\end{align}
For every task map $G$,
$KG=\overline K(qG)$, so the joint spectrum depends only on the equivalence classes of the terminal task functions modulo $\ker K$.
\end{proposition}

\paragraph{Proof of \Cref{prop:quotient}.}\label{app:proof-quotient}

Define
$\overline K([h])=Kh$.  The definition is independent of the representative because two representatives differ by an element of $\ker K$.  It is injective by construction and satisfies $K=\overline Kq$.
For the quotient norm
$\lVert[h]\rVert=\inf_{v\in\ker K}\lVert h+v\rVert$, boundedness of $K$ gives
\begin{align*}
\lVert\overline K[h]\rVert
=
\lVert K(h+v)\rVert
\le
\lVert K\rVert_{\opn}\lVert h+v\rVert
\end{align*}
for every $v\in\ker K$.  Taking the infimum shows
$\lVert\overline K\rVert_{\opn}\le\lVert K\rVert_{\opn}$.  Conversely,
$\lVert qh\rVert\le\lVert h\rVert$, so
$\lVert K\rVert_{\opn}\le\lVert\overline K\rVert_{\opn}$.  The norms are equal.  The identity for $KG$ follows by composition.
\hfill $\blacksquare$

\subsection{Coordinate changes on linear observables}\label{app:linear-coordinate-realization}

For example, on the space $\mathcal V=\{z\mapsto c^\top z:c\in\mathbb R^d\}$ of linear observables, every $R\in\operatorname{GL}(\mathcal V)$ is induced by an invertible hidden-coordinate map.  Under the coefficient identification, choose $B=R^\top$; then the Koopman pullback $K_B$ acts as $R$, and
\begin{align*}
K_{B^{-1}\circ\Phi}\,K_B G
=
K_\Phi G.
\end{align*}
Linear observables therefore give a concrete Koopman realization of every coordinate change in the theorem.

\FloatBarrier

\section{Why separate representation and task bounds can be loose}\label{sec:no-go}

Neural generalization bounds often use parameter norms, products of layer norms or compression~\citep{Neyshabur2015,bartlett2017spectrally,golowich2018size,arora2018compression}. Equivalent parameterizations can change such summaries while preserving predictions~\citep{DinhEtAl2017SharpMinima}. Path norms, Fisher--Rao geometry and path metrics address this dependence in different ways~\citep{NeyshaburSalakhutdinovSrebro2015PathSGD,LiangPoggioRakhlinStokes2019,Gonon2025PathMetrics}. The following constructions isolate the effect of separating a shared operator from its task functions, and of multiplying layer norms when transformations cancel.

Define
\begin{align*}
\Ojoint(K,G)=\lVert KG\Dn^{-1/2}\rVert_*,
\qquad
\Osep(K,G)=\lVert K\rVert_{\opn}\lVert G\Dn^{-1/2}\rVert_*.
\end{align*}
The ideal property gives $\Ojoint\le\Osep$, and \Cref{thm:optimal-refactor} identifies the joint value as the infimum over invertible changes of intermediate coordinates. For an arbitrary factorization, the gap can be unbounded. The examples below realize this gap with composition operators acting on linear functions.

\subsection{Equivalent factorizations}\label{app:equivalent-factorizations}

\begin{theorem}[Equivalent factorizations can make the separated bound arbitrarily loose]\label{thm:no-go}
Let $K:\mathcal H_L\to\mathcal H_0$ be bounded, let $G:\mathbb R^\Task\to\mathcal H_L$, and let $R:\mathcal H_L\to\mathcal H_L$ be bounded and invertible.  Set
\begin{align*}
K_R
=
KR^{-1},
\qquad
G_R
=
RG.
\end{align*}
Then
\begin{align}
K_RG_R
=
KG,
\qquad
\Ojoint\left(K_R,G_R\right)
=
\Ojoint\left(K,G\right).
\label{eq:factorization-invariance}
\end{align}
There is no universal constant $C$ such that
\begin{align*}
\Osep\left(K,G\right)
\le
C\Ojoint\left(K,G\right)
\end{align*}
for all finite-dimensional Hilbert spaces, invertible $K$, and rank-one $G$.  More precisely, for every $a\ge1$ there are equivalent factorizations of one fixed predictor map for which
\begin{align}
\Ojoint
=
1,
\qquad
\Osep
=
a^2.
\label{eq:unbounded-gap}
\end{align}
\end{theorem}

\paragraph{Proof of \Cref{thm:no-go}.}\label{app:proof-no-go}

The first identity follows by direct multiplication:
$K_RG_R=KR^{-1}RG=KG$.  The joint geometry depends only on this product, so \Cref{eq:factorization-invariance} follows.

For the unbounded gap, take
$\mathcal H_L=\mathcal H_0=\mathbb R^2$, $\Task=1$, and $\Dn=[1]$.  Let the fixed predictor map be
$Fe_1=e_2$.  Start from $K=I_2$ and $Ge_1=e_2$.  For $a\ge1$, define
\begin{align*}
R_a
=
\operatorname{diag}\left(a^{-1},a\right),
\qquad
K_a
=
R_a^{-1}
=
\operatorname{diag}\left(a,a^{-1}\right),
\qquad
G_a
=
R_aG.
\end{align*}
Then $G_ae_1=ae_2$ and
$K_aG_ae_1=e_2$, so the predictor is unchanged and
$\Ojoint(K_a,G_a)=1$.  On the other hand,
$\Osep(K_a,G_a)=\lVert K_a\rVert_{\opn}\lVert G_a\rVert_*=a^2$.
Choosing $a>\sqrt C$ rules out every universal reverse constant.
\hfill $\blacksquare$

\subsection{Relative orientation}\label{app:relative-orientation}

Equivalent factorizations can thus preserve every prediction while making the separated product arbitrarily large.  The next construction isolates a second source of information loss: even when the singular values of each factor are fixed, their relative directions can change the composed norm.

\begin{proposition}[Separate singular values do not determine the composed norm]\label{prop:orientation}
Let $\mathcal H_L=\mathcal H_0=\mathbb R^2$, let $\Task=1$, and let $\Dn=[1]$.  For every $a>1$, there exist one representation map $K_a$ and two rank-one task maps $G_0,G_{\pi/2}$ such that
\begin{itemize}[leftmargin=1.35em,itemsep=.1em,topsep=.1em]
\item $G_0$ and $G_{\pi/2}$ have the same singular values;
\item the singular values of $K_a$ are the same in both constructions;
\item the separated products are equal;
\item the joint geometries satisfy
\begin{align}
\frac{\Ojoint\left(K_a,G_0\right)}{\Ojoint\left(K_a,G_{\pi/2}\right)}
=
a^2.
\label{eq:orientation-ratio}
\end{align}
\end{itemize}
More generally, if a unit task direction forms angle $\vartheta$ with the most-expanded singular direction of
$K_a=\operatorname{diag}\left(a,a^{-1}\right)$, its joint norm is
\begin{align}
\left(
a^2\cos^2\vartheta
+
a^{-2}\sin^2\vartheta
\right)^{1/2},
\label{eq:orientation-formula}
\end{align}
while the separated product remains $a$.
\end{proposition}

\begin{corollary}[No characterization from separate singular values alone]\label{cor:no-marginal-spectrum}
There is no function of the singular values of $K$ and $G\Dn^{-1/2}$ alone that is uniformly equivalent, up to a finite multiplicative constant, to
$\lVert KG\Dn^{-1/2}\rVert_*$.  In particular, separate spectral summaries cannot determine the weighted norm of the composed predictor map without information about relative singular directions.
\end{corollary}

\paragraph{Proof of \Cref{prop:orientation,cor:no-marginal-spectrum}.}\label{app:proof-orientation}

Let
$K_a=\operatorname{diag}\left(a,a^{-1}\right)$,
$G_0e_1=e_1$, and
$G_{\pi/2}e_1=e_2$.  Both task maps have the single nonzero singular value one, and $K_a$ is the same in both constructions.  Hence both separated products equal $a$.  Their joint norms are
$\lVert K_aG_0\rVert_*=a$ and
$\lVert K_aG_{\pi/2}\rVert_*=a^{-1}$, proving \Cref{eq:orientation-ratio}.
For
$v_\vartheta=\cos\vartheta\,e_1+\sin\vartheta\,e_2$,
\begin{align*}
\lVert K_av_\vartheta\rVert_2^2
=
a^2\cos^2\vartheta+a^{-2}\sin^2\vartheta,
\end{align*}
which proves \Cref{eq:orientation-formula}.

Suppose a statistic depending only on the two marginal singular spectra were within a factor $C$ of the joint geometry.  The two constructions above have identical inputs to that statistic, so their joint geometries could differ by at most $C^2$.  Equation~\eqref{eq:orientation-ratio} makes the ratio $a^2$, and $a$ is arbitrary.  No finite $C$ can work.
\hfill $\blacksquare$

\subsection{Cancellation across layers}\label{app:layer-cancellation}

\begin{theorem}[Exponential gap for cancelling deep linear layers]\label{thm:depth-gap}
Let $a>1$, let $m\ge1$ be an integer, and let
$A_a=\operatorname{diag}\left(a,a^{-1}\right)$ on $\mathbb R^2$.  Consider the Hilbert space of linear observables
$\mathcal V=\{x\mapsto c^\top x:c\in\mathbb R^2\}$ with norm $\lVert c\rVert_2$.  For the linear layer $x\mapsto Wx$, its Koopman pullback acts on coefficients as $c\mapsto W^\top c$ and therefore has operator norm $\lVert W\rVert_{\opn}$.  A depth-$2m$ network formed from $m$ adjacent pairs $A_a^{-1}A_a$ represents the identity map and induces the identity pullback on $\mathcal V$, but
\begin{align}
\prod_{\ell=1}^{2m}
\left\lVert
K_\ell
\right\rVert_{\opn}
=
a^{2m}.
\label{eq:depth-gap}
\end{align}
Consequently, for every nonzero task map $G$ into $\mathcal V$, a bound that replaces the end-to-end geometry by the product of layer pullback norms times
$\lVert G\Dn^{-1/2}\rVert_*$ can be exponentially larger in depth than the invariant joint geometry.
\end{theorem}

\paragraph{Proof of \Cref{thm:depth-gap}.}\label{app:proof-depth}

Identify $g_c(x)=c^\top x\in\mathcal V$ with its coefficient vector $c$.  For a linear map $W$, the Koopman pullback satisfies
\begin{align*}
K_Wg_c(x)
=
g_c(Wx)
=
(W^\top c)^\top x,
\end{align*}
so its matrix on $\mathcal V$ is $W^\top$ and $\lVert K_W\rVert_{\opn}=\lVert W\rVert_{\opn}$.  Both $A_a$ and $A_a^{-1}$ have norm $a$.  Every adjacent pair of network maps is the identity, hence the complete network and its pullback on $\mathcal V$ are the identity.  Multiplying the $2m$ individual pullback norms gives $a^{2m}$, which is \Cref{eq:depth-gap}.  For any task map $G$ into $\mathcal V$, the joint end-to-end map is $G$, whereas the layerwise product relaxation multiplies its task norm by $a^{2m}$.
\hfill $\blacksquare$

The examples distinguish changes of coordinates, relative alignment and cancellation across layers. The composed norm accounts for each interaction; the corresponding product of separate norms does not. \Cref{sec:estimated-feature-geometry} studies how to estimate the Gram matrix used by the composed penalty.

\subsection{Plots of the deterministic constructions}\label{app:analytic-constructions}

\begin{figure}[!htbp]
\centering
\begin{minipage}[t]{0.49\linewidth}
\centering\vspace{0pt}\textbf{(a)}\par\smallskip
\includegraphics[width=\linewidth]{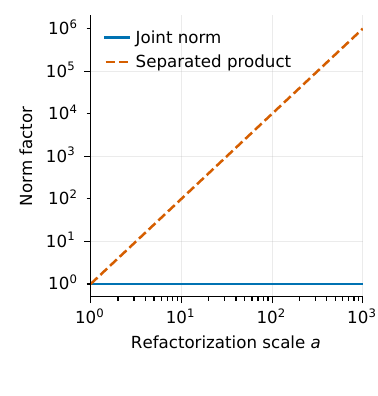}
\end{minipage}\hfill
\begin{minipage}[t]{0.49\linewidth}
\centering\vspace{0pt}\textbf{(b)}\par\smallskip
\includegraphics[width=\linewidth]{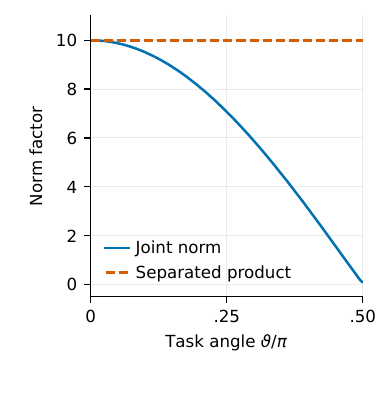}
\end{minipage}
\par\medskip
\begin{minipage}[t]{0.49\linewidth}
\centering\vspace{0pt}\textbf{(c)}\par\smallskip
\includegraphics[width=\linewidth]{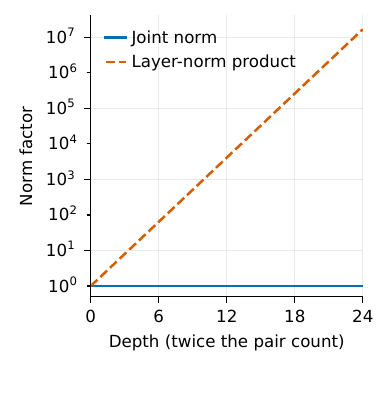}
\end{minipage}
\caption{Deterministic mechanism checks.  (a) Equivalent factorizations of one predictor keep joint geometry fixed while the separated product grows.  (b) Fixed marginal spectra and a fixed separated product do not determine joint geometry.  (c) Cancelling operator pairs keep the composed map fixed while layer-norm products grow exponentially}
\label{fig:analytic-gaps}
\end{figure}

The vector figures are generated with Matplotlib by the accompanying \path{generate_revised_figures.py}. \Cref{fig:analytic-gaps} plots the exact formulas below. Its plotting grids are provided as \path{intermediate_refactorization_gap.csv}, \path{alignment_gap.csv} and \path{depth_cancellation.csv}.

\paragraph{Refactorization panel.}
For $a\in[1,10^3]$, the fixed predictor map in the proof of \Cref{thm:no-go} has joint geometry one and separated product $a^2$.

\paragraph{Alignment panel.}
The panel uses $a=10$ and plots the exact expression in \Cref{eq:orientation-formula} for $\vartheta\in[0,\pi/2]$, together with the constant separated value $a$.

\paragraph{Depth panel.}
The panel uses $a=2$ and $m=0,\ldots,12$ cancelling pairs.  The composed map is the identity, whereas the product of layer norms is $2^{2m}$.

\section{Task Gram matrices from function values and derivatives}\label{sec:pullback}

We first give the finite-feature construction used in \Cref{sec:estimated-feature-geometry}. The later composition formulas impose additional assumptions on the shared map and describe further realizations of the task Gram.

\subsection{Finite features with values and derivatives}\label{app:derivative-features}

For differentiable $b=(b_1,\ldots,b_p)^\top$ on
$\Omega\subseteq\mathbb R^d$, let
$J_b(x)\in\mathbb R^{p\times d}$ have rows $\nabla b_j(x)^\top$.
The choice
\begin{align}
W(x)=b(x)b(x)^\top+J_b(x)J_b(x)^\top
\label{eq:derivative-feature-contribution}
\end{align}
gives a Gram matrix by averaging products of feature values and derivatives under $\mu$. For
$f_t(x)=b(x)^\top A e_t$, the task Gram estimator in
\Cref{eq:empirical-pullback-gram} is exactly $A^\top\widehat Q A$.
To use $J_Q$ as $\mJ(F_A)$, the feature inner product must be the one
whose Gram is $Q$.  For example, if $\Omega$ is open, $\mu$ has full
support on $\Omega$, and the features are continuously differentiable
and their values and derivatives have finite second moments, their finite span with this
energy inner product is an RKHS after redundant coefficients are
identified.  Its kernel is
$k_Q(x,x')=b(x)^\top Q^\dagger b(x')$; the proof follows below.
This finite-dimensional construction is available in every input
dimension.  A bounded kernel diagonal, when needed for
\Cref{eq:kernel-bound}, remains an additional condition.

\paragraph{Finite-span RKHS construction.}\label{app:proof-derivative-features}

For $b:\Omega\to\mathbb R^p$ as above,
$W=b b^\top+J_bJ_b^\top$ is positive semidefinite and measurable.
Its expected trace is finite under the stated moments.  If
$f_t=b^\top Ae_t$, then $\nabla f_t=J_b^\top Ae_t$; hence
\begin{align*}
f_t(x)f_s(x)+\langle\nabla f_t(x),\nabla f_s(x)\rangle
=e_t^\top A^\top W(x)Ae_s.
\end{align*}
Averaging proves $\widehat S=A^\top\widehat Q A$.

Now assume $\Omega$ is open, $\mu$ has full support there, and the
features are continuously differentiable.  For $a\in\mathbb R^p$,
write $f_a=b^\top a$.  Then
\begin{align*}
a^\top Qa
=\mathbb E\{f_a(U)^2+\lVert\nabla f_a(U)\rVert_2^2\}.
\end{align*}
If $a\in\ker Q$, then $f_a=0$ $\mu$-almost everywhere.
Continuity and full support imply $f_a(x)=0$ for every $x\in\Omega$:
a nonzero value would give an open neighborhood of positive
$\mu$-measure where $|f_a|$ is bounded away from zero.  Conversely,
if $f_a$ vanishes everywhere on the open domain, its gradient also
vanishes there, and $a^\top Qa=0$.  Thus the kernel of the map
$a\mapsto f_a$ is precisely $\ker Q$.

The quotient $\mathbb R^p/\ker Q$ consequently identifies with the
actual function span $\mathcal H_b=\{f_a:a\in\mathbb R^p\}$, and
\begin{align*}
\langle f_a,f_c\rangle_{\mathcal H_b}=a^\top Qc
\end{align*}
is a well-defined inner product.  This finite-dimensional space is
complete.  Since $a^\top b(x)=0$ for every $a\in\ker Q$,
$b(x)\in\operatorname{range}(Q)$ for each $x$.  Put
$k_x=f_{Q^\dagger b(x)}$.  Then
\begin{align*}
\langle f_a,k_x\rangle_{\mathcal H_b}
&=a^\top QQ^\dagger b(x)=a^\top b(x)=f_a(x),
\\
\lVert k_x\rVert_{\mathcal H_b}^2
&=b(x)^\top Q^\dagger b(x)<\infty.
\end{align*}
Cauchy--Schwarz makes every point evaluation continuous, proving the
RKHS assertion and the kernel formula.  This does not assert bounded
evaluation in the full infinite-dimensional $H^1(\Omega)$ space.
If full support or the pointwise compatibility with $\ker Q$ fails,
the energy quotient still exists algebraically, but it need not identify
functions with well-defined evaluation on the requested input domain.
\hfill $\blacksquare$

\subsection{Estimating the Gram of fixed predictors}\label{app:empirical-derivative-gram}

Fix differentiable predictors with measurable values and gradients, or condition on the data used to fit them. For a chosen probability measure $\mu$, assume
\begin{align*}
\mathbb E_{U\sim\mu}\bigl[f_t(U)^2+\lVert\nabla f_t(U)\rVert_2^2\bigr]<\infty
\qquad(t\in[\Task]).
\end{align*}
Define the distribution-weighted pullback Gram by
\begin{align*}
S_{ts}^{\mathrm{pb},\mu}
=
\mathbb E_{U\sim\mu}
\left[
 f_t(U)f_s(U)
+
\left\langle
\nabla f_t(U),
\nabla f_s(U)
\right\rangle
\right].
\end{align*}
Given $U_1,\ldots,U_m$ i.i.d. from $\mu$, independently of the fitted predictors, its empirical counterpart is
\begin{align}
\widehat S_{ts}^{\mathrm{pb},\mu}
=
\frac{1}{m}
\sum_{j=1}^{m}
\left[
 f_t(U_j)f_s(U_j)
+
\left\langle
\nabla f_t(U_j),
\nabla f_s(U_j)
\right\rangle
\right].
\label{eq:empirical-pullback-gram}
\end{align}
Equation~\eqref{eq:empirical-pullback-gram} is conditionally unbiased for the probability-weighted Gram $S^{\mathrm{pb},\mu}$; the moment assumption and Cauchy--Schwarz ensure integrability of every entry.  If $\mu$ is normalized Lebesgue measure on $\Omega_0$, then
$|\Omega_0|S^{\mathrm{pb},\mu}$ is the unnormalized $H^1(\Omega_0)$ Gram and $|\Omega_0|\widehat S^{\mathrm{pb},\mu}$ is its unbiased empirical estimator.  For a sampling density $p$, recovery of the Lebesgue Gram requires weights $1/p(U_j)$ and $p>0$ almost everywhere on the relevant energy support.  The regularizer
$\operatorname{Tr}\left[\left(\Dn^{-1/2}\widehat S^{\mathrm{pb},\mu}\Dn^{-1/2}\right)^{1/2}\right]$
depends only on the predictors and their input derivatives, and is unchanged by hidden-coordinate transformations that preserve them. The square-root trace is nonlinear, so the entrywise unbiasedness of $\widehat S^{\mathrm{pb},\mu}$ does not imply unbiasedness of this regularizer. \Cref{sec:estimated-feature-geometry} gives a uniform approximation guarantee when the predictors lie in a fixed finite feature span and the reference contributions satisfy a leverage bound.

\subsection{Finite-to-one composition in Sobolev spaces}\label{app:finite-to-one-sobolev}

The $H^1$ inner product measures both function values and first derivatives. Under composition, its first term changes with volume and multiplicity, while its derivative term also depends on directional distortion. Boundedness criteria for Sobolev composition operators describe these effects~\citep{HenclKoskela2008FiniteDistortion,MenovschikovUkhlov2021QMappings,OlivaPrats2017Composition}; further results cover higher-order and geometric conditions~\citep{IkedaIshikawaTaniguchi2024,Ukhlov2024GeometricCharacterizations}. Here the chain rule and area formula give the Gram matrix of the composed task functions when the shared map is finite-to-one and Lipschitz. Let $\Omega_0,\Omega_L\subset\mathbb R^d$ be bounded Lipschitz open sets and let
$\Phi:\Omega_0\to\Omega_L$ be Lipschitz.  Assume that
\begin{itemize}[leftmargin=1.35em,itemsep=.1em,topsep=.1em]
\item $\Phi$ has finite multiplicity almost everywhere;
\item its Jacobian
$\Jphi(x)=\left\lvert\det D\Phi(x)\right\rvert$
is positive almost everywhere.
\end{itemize}
We use the first-order Sobolev inner product
\begin{align*}
\left\langle u,v\right\rangle_{H^1(\Omega)}
=
\int_{\Omega}u(x)v(x)\,dx
+
\int_{\Omega}\nabla u(x)^\top\nabla v(x)\,dx.
\end{align*}
For a measurable matrix field $M$, $\lVert M\rVert_{L^\infty(\opn)}$ denotes the essential supremum of its matrix operator norm.

For almost every $y\in\Omega_L$, define
\begin{align}
\mphi(y)
&=
\sum_{x\in\Phi^{-1}(y)}
\frac{1}{\Jphi(x)},
\label{eq:multiplicity-density}
\\
\Mphi(y)
&=
\sum_{x\in\Phi^{-1}(y)}
\frac{D\Phi(x)D\Phi(x)^\top}{\Jphi(x)}.
\label{eq:distortion-tensor}
\end{align}
Empty sums are zero.  The scalar $\mphi$ records multiplicity and volume change; the positive-semidefinite matrix $\Mphi$ records directional distortion accumulated over all preimages.

\begin{theorem}[First-order Sobolev formula for finite-to-one maps]\label{thm:pullback-formula}
For every $g,h\in C^1\left(\overline{\Omega_L}\right)$,
\begin{align}
\left\langle
g\circ\Phi,
h\circ\Phi
\right\rangle_{H^1\left(\Omega_0\right)}
&=
\int_{\Omega_L}
\mphi(y)g(y)h(y)\,dy
\notag\\
&\quad+
\int_{\Omega_L}
\nabla g(y)^\top
\Mphi(y)
\nabla h(y)\,dy.
\label{eq:pullback-formula}
\end{align}
If
\begin{align*}
\mphi\in L^\infty\left(\Omega_L\right),
\qquad
\left\lVert\Mphi\right\rVert_{L^\infty(\opn)}<\infty,
\end{align*}
then composition extends uniquely to a bounded operator
$K_\Phi:H^1\left(\Omega_L\right)\to H^1\left(\Omega_0\right)$ satisfying
\begin{align}
\left\lVert
K_\Phi
\right\rVert_{\opn}^2
\le
\max\left\{
\left\lVert\mphi\right\rVert_\infty,
\left\lVert\Mphi\right\rVert_{L^\infty(\opn)}
\right\}.
\label{eq:pullback-operator-bound}
\end{align}
\end{theorem}

Unlike an injective change-of-variables formula, \Cref{eq:pullback-formula} sums over every preimage.  It therefore distinguishes a one-to-one representation from a folded representation with the same local Jacobian on each branch.  For example, take $\Omega_0=(-1/2,1/2)$, $\Omega_L=(-1,1)$, and $\Phi(x)=2|x|$.  For almost every $y\in(0,1)$ there are two preimages with Jacobian magnitude two, so $\mphi(y)=1$ and, in one dimension, $\Mphi(y)=4$; both fields vanish on $(-1,0)$.  Thus the $L^2$ energy on the realized image is unchanged while the derivative energy there is multiplied by four, making multiplicity and directional stretching explicit.

\begin{corollary}[Task geometry after the shared map]\label{cor:pullback-spectrum}
Assume the hypotheses and the two $L^\infty$ boundedness conditions of \Cref{thm:pullback-formula}.  By density of $C^1(\overline{\Omega_L})$ in $H^1(\Omega_L)$ and continuity of the bounded pullback, the bilinear identity in \Cref{eq:pullback-formula} extends to every pair in $H^1(\Omega_L)$.  Let $g_1,\ldots,g_\Task\in H^1\left(\Omega_L\right)$ and let
$F_\Phi=K_\Phi G_{\bg}$.  Its end-to-end Gram matrix has entries
\begin{align}
\left(S_{\Phi,\bg}^{\mathrm{pb}}\right)_{ts}
&=
\int_{\Omega_L}
\mphi(y)g_t(y)g_s(y)\,dy
\notag\\
&\quad+
\int_{\Omega_L}
\nabla g_t(y)^\top
\Mphi(y)
\nabla g_s(y)\,dy.
\label{eq:pullback-task-gram}
\end{align}
Consequently,
\begin{align}
\left\lVert
F_\Phi\Dn^{-1/2}
\right\rVert_*
=
\operatorname{Tr}
\left[
\left(
\Dn^{-1/2}
S_{\Phi,\bg}^{\mathrm{pb}}
\Dn^{-1/2}
\right)^{1/2}
\right].
\label{eq:pullback-joint-spectrum}
\end{align}
In one dimension on a bounded interval, $H^1$ has bounded point evaluation, so this geometry can be inserted directly into \Cref{thm:exact-rademacher}.  In dimensions $d\ge2$, \Cref{eq:pullback-task-gram} remains an exact Sobolev energy identity, but a direct pointwise Rademacher bound requires a smoother RKHS or another evaluation-continuous function space.
\end{corollary}

\Cref{app:cm-specialization} develops the anchored one-dimensional case, where the derivative energy alone defines an evaluation-continuous RKHS.

\paragraph{Proof of \Cref{thm:pullback-formula,cor:pullback-spectrum}.}\label{app:proof-pullback}

For smooth $g$, the chain rule gives
$\nabla(g\circ\Phi)(x)=D\Phi(x)^\top\nabla g(\Phi(x))$ almost everywhere.  Taking $h=g$ first gives
\begin{align}
\left\langle
g\circ\Phi,h\circ\Phi
\right\rangle_{H^1(\Omega_0)}
&=
\int_{\Omega_0}
g\left(\Phi(x)\right)
h\left(\Phi(x)\right)\,dx
\notag\\
&\quad+
\int_{\Omega_0}
\nabla g\left(\Phi(x)\right)^\top
D\Phi(x)D\Phi(x)^\top
\nabla h\left(\Phi(x)\right)\,dx.
\label{eq:proof-pullback-before-area}
\end{align}
The area formula for Lipschitz maps states that, for every nonnegative measurable $r$,
\begin{align*}
\int_{\Omega_0}
r(x)\Jphi(x)\,dx
=
\int_{\Omega_L}
\sum_{x\in\Phi^{-1}(y)}r(x)\,dy.
\end{align*}
Because $\Jphi>0$ almost everywhere, apply the formula to the two nonnegative quadratic integrands divided by $\Jphi(x)$.  This proves \Cref{eq:pullback-formula} for $h=g$.  The real bilinear identity follows by polarization, applied to $g+h$ and $g-h$.

Since $\Mphi(y)$ is positive semidefinite,
\begin{align*}
\left\lVert g\circ\Phi\right\rVert_{H^1(\Omega_0)}^2
&\le
\left\lVert\mphi\right\rVert_\infty
\left\lVert g\right\rVert_{L^2(\Omega_L)}^2
+
\left\lVert\Mphi\right\rVert_{L^\infty(\opn)}
\left\lVert\nabla g\right\rVert_{L^2(\Omega_L)}^2
\notag\\
&\le
\max\left\{
\left\lVert\mphi\right\rVert_\infty,
\left\lVert\Mphi\right\rVert_{L^\infty(\opn)}
\right\}
\left\lVert g\right\rVert_{H^1(\Omega_L)}^2.
\end{align*}
Density of smooth functions on bounded Lipschitz open sets gives the unique bounded extension and \Cref{eq:pullback-operator-bound}.  The $L^2$ part of the same estimate identifies the extension with ordinary composition almost everywhere.

Apply \Cref{eq:pullback-formula} with $g=g_t$ and $h=g_s$.  The resulting matrix is
$G_{\bg}^*K_\Phi^*K_\Phi G_{\bg}$, proving \Cref{eq:pullback-task-gram}.  Equation~\eqref{eq:pullback-joint-spectrum} follows from the singular-value identity used in \Cref{thm:end-covariance}.
\hfill $\blacksquare$

\subsection{The Cameron--Martin space on an interval}\label{app:cm-specialization}

On an interval, the Cameron--Martin inner product depends only on derivatives and gives a simpler expression for the task Gram matrix. Point evaluations are continuous in this space. Let $I_0,I_L$ be bounded intervals containing the origin and let
\begin{align*}
\mathcal H_{\mathrm{CM}}(I)
=
\left\{
 g:g(0)=0,\ g\text{ absolutely continuous},\ g'\in L^2(I)
\right\},
\\
\lVert g\rVert_{\mathrm{CM}}^2
=
\int_I\left\lvert g'(u)\right\rvert^2\,du.
\end{align*}

\Needspace{8\baselineskip}
\begin{corollary}[Finite-to-one Cameron--Martin pullback]\label{cor:cm-pullback}
Let $\phi:I_0\to I_L$ be Lipschitz and finite-to-one, assume
$\phi(0)=0$ and $\left\lvert\phi'(x)\right\rvert>0$ almost everywhere, and define
\begin{align}
q_\phi(y)
=
\sum_{x\in\phi^{-1}(y)}
\left\lvert\phi'(x)\right\rvert.
\label{eq:cm-weight}
\end{align}
For all $g,h\in C^1(\overline I_L)$ satisfying $g(0)=h(0)=0$,
\begin{align}
\left\langle
 g\circ\phi,
 h\circ\phi
\right\rangle_{\mathrm{CM}}
=
\int_{I_L}
q_\phi(y)g'(y)h'(y)\,dy.
\label{eq:cm-pullback}
\end{align}
If $q_\phi\in L^\infty(I_L)$, composition extends uniquely to a bounded operator
$K_\phi:\mathcal H_{\mathrm{CM}}(I_L)\to\mathcal H_{\mathrm{CM}}(I_0)$, the identity in \Cref{eq:cm-pullback} holds for all $g,h$ in the Cameron--Martin space, and
$\lVert K_\phi\rVert_{\opn}^2\le\lVert q_\phi\rVert_\infty$.  The corresponding multitask Gram matrix is obtained from \Cref{eq:cm-pullback}, and its sample-size-normalized nuclear spectrum enters \Cref{thm:exact-rademacher} directly.
\end{corollary}

\paragraph{Proof of \Cref{cor:cm-pullback}.}\label{app:proof-cm}

For anchored $C^1$ functions, the anchoring condition gives $(g\circ\phi)(0)=g(0)=0$, and the one-dimensional chain rule yields
$(g\circ\phi)'(x)=g'(\phi(x))\phi'(x)$ almost everywhere.  Hence
\begin{align*}
\left\langle
 g\circ\phi,
 h\circ\phi
\right\rangle_{\mathrm{CM}}
=
\int_{I_0}
g'\left(\phi(x)\right)
h'\left(\phi(x)\right)
\phi'(x)^2\,dx.
\end{align*}
For $h=g$, apply the one-dimensional area formula after dividing by
$\left\lvert\phi'(x)\right\rvert$.  Since
$\phi'(x)^2/\left\lvert\phi'(x)\right\rvert=\left\lvert\phi'(x)\right\rvert$, this gives the quadratic identity on the anchored $C^1$ core; the bilinear form in \Cref{eq:cm-pullback} follows by polarization.  If $q_\phi\in L^\infty$, taking $g=h$ gives
\begin{align*}
\lVert g\circ\phi\rVert_{\mathrm{CM}}^2
\le
\lVert q_\phi\rVert_\infty
\lVert g\rVert_{\mathrm{CM}}^2.
\end{align*}
Anchored $C^1$ functions are dense in $\mathcal H_{\mathrm{CM}}(I_L)$, so composition extends uniquely and the bilinear identity passes to the limit for arbitrary Cameron--Martin functions.
\hfill $\blacksquare$

\FloatBarrier

\section{Controlled composition and Gram-estimation studies}\label{app:composition-pilot}

The controlled composition study has a separate protocol and outputs from the task-count experiments in \Cref{app:legacy-experiments}. The machine-readable protocol, implementation and implementation-check report were bound by SHA-256 before the first predictive fit. Every declared cell and seed is retained. The accompanying code-and-data package contains the raw inputs, feature matrices, responses, noise, true coefficients, every fitted regularization path, independent validation losses, selected coefficients, dual certificates, reference inputs and empirical Grams.

\subsection{Composition and prediction}\label{app:composition-prediction}

\paragraph{Data and truth.}
For each of $16$ seeds, task inputs are independent uniform draws on $[-\pi,\pi]$ and response noise is independent $N(0,1)$. The same inputs, noise and orthonormal task factor $V\in\mathbb R^{6\times2}$ are used across all paired cells. Set $F^\star=U\operatorname{diag}(s_1,s_2)V^\top$, where $(s_1,s_2)$ is proportional to $(1,0.6)$ and normalized so that $\|F^\star\|_{\HS}^2/6=1$. With coordinates numbered from one, the columns of $U$ are $(e_1,e_2)$, $(e_7,e_8)$ and $((e_1+e_8)/\sqrt2,(e_2+e_7)/\sqrt2)$ for signals in the most amplified, least amplified and mixed directions, respectively. The entries of $K$ form a geometric progression from $1$ to $1/\rho$. Consequently the truth and observations do not change with $\rho$ within an alignment cell; only the representation and the separated penalty change.

\paragraph{Optimization and selection.}
Every method uses $17$ parameters, geometrically spaced from its own zero-solution threshold $\lambda_{\max}$ to $10^{-4}\lambda_{\max}$. The selected candidate minimizes task-average validation MSE; an exact tie selects the larger parameter. There is no refitting after selection. For nuclear-penalized variable $B$, write the prediction coefficients as $F=SB$, where $S=\sqrt nKQ^{-1/2}$ for the composed implementation, $S=\sqrt nK$ for the separated method, and $S=\sqrt n I$ for the orthonormalized implementation. Each objective is convex and is solved by accelerated proximal gradient with singular-value thresholding and adaptive restart.

For a candidate fit, define the residual $r_t=X_tSb_t-y_t$ and the scaled residual $u_t=r_t/(6n)$. A feasible dual point is obtained by multiplying each $u_t$ by $\alpha=\min\{1,\lambda/\|[S^\top X_t^\top u_t]_{t=1}^6\|_{\opn}\}$, with $\alpha=1$ when the denominator is zero. The feasible dual value is
\[
-\frac{6n}{2}\sum_{t=1}^6\|\alpha u_t\|_2^2
-\sum_{t=1}^6y_t^\top\alpha u_t.
\]
The relative gap is the primal objective minus this value, divided by the maximum of one and the absolute primal objective. All $7{,}344$ candidates converge with relative gap at most $9.983\times10^{-8}$. Among $432$ selected models, one selects the smallest declared parameter and none selects the largest. No candidate or seed is excluded. Implementation checks compare with a closed-form singular-value-thresholding problem and independently reconstruct the residual dual objective.

\paragraph{Effects and controls.}
The paired gain is separated population excess minus composed population excess. Table~\ref{tab:composition-pilot} includes every declared cell. Intervals are two-sided Student-$t$ intervals for $16$ independent paired-seed effects within a cell, reported descriptively for this study without multiplicity adjustment. The largest coefficient-matrix difference is zero both for the isotropic null and for the composed--orthonormalized comparison. For an additional fixed nonorthogonal hidden-coordinate change, the largest predictor difference is $2.681\times10^{-16}$ and the largest composed-penalty difference is $2.221\times10^{-16}$.

\begin{table}[!htbp]
\centering\small
\begin{tabular}{rlrrr}
\toprule
$\rho$ & Truth directions & Composed & Separated & Gain [$95\%$ interval]\\
\midrule
1 & Largest & 0.13503 & 0.13503 & $0\ [0,0]$\\
1 & Smallest & 0.13315 & 0.13315 & $0\ [0,0]$\\
1 & Mixed & 0.13748 & 0.13748 & $0\ [0,0]$\\
4 & Largest & 0.13503 & 0.08591 & $-0.04912\ [-0.06114,-0.03710]$\\
4 & Smallest & 0.13315 & 0.18519 & $0.05204\ [0.04039,0.06369]$\\
4 & Mixed & 0.13748 & 0.17608 & $0.03860\ [0.02971,0.04749]$\\
16 & Largest & 0.13503 & 0.06674 & $-0.06829\ [-0.08333,-0.05326]$\\
16 & Smallest & 0.13315 & 0.20946 & $0.07631\ [0.05902,0.09360]$\\
16 & Mixed & 0.13748 & 0.20928 & $0.07180\ [0.05607,0.08752]$\\
\bottomrule
\end{tabular}
\caption{Exact population excess and paired gains in all $9$ prespecified composition cells. Each mean uses $16$ paired seeds.}
\label{tab:composition-pilot}
\end{table}

\FloatBarrier
\subsection{Estimating the feature Gram matrix}\label{app:gram-study}
For active rank $r\in\{4,8\}$, use the first $r$ paired trigonometric coordinates and set the remaining shared-feature coefficients to zero. The nonzero singular values of $K$ range from $1$ to $1/\rho$ for $\rho\in\{1,16\}$. Each of $32$ independent replicates draws $1{,}024$ reference inputs and uses its prefixes of lengths $16,32,64,128,256,512,1{,}024$. The exact and empirical Grams are $Q=K^\top K$ and $\widehat Q=K^\top(m^{-1}\sum_j\varphi(U_j)\varphi(U_j)^\top)K$. On the range of $Q$, we record the spectral error $\|Q^{\dagger/2}\widehat Q Q^{\dagger/2}-I_{\operatorname{ran}Q}\|_{\opn}$. All nullspace leakages are zero. The sum of squared active trigonometric features is $r$ at every input, so the whitened leverage is exactly $r$, independently of $\rho$. The recorded smallest and largest whitened eigenvalues give the extremal possible ratios of empirical to exact nuclear penalties. Rank-one task heads in the corresponding eigenvector directions attain these ratios; $32$ additional Gaussian heads are recorded in each case. This study concerns the $L^2$ feature Gram and does not evaluate derivative-based Sobolev Grams or representation training.

\FloatBarrier

\section{SARCOS application protocol and numerical records}
\label{app:sarcos-application}

\paragraph{Data, preparation and observation overlap.}
We use the official GPML SARCOS files \texttt{sarcos\_inv.mat}
($44{,}484\times28$) and \texttt{sarcos\_inv\_test.mat}
($4{,}449\times28$) \citep{SARCOSGPMLData}.
The first 21 columns contain inputs and the final seven contain joint
 torques.  One fixed random preparation set of 1,024 official training
rows determines each input and output mean and standard deviation
(using divisor 1,024).  These rows are excluded from all supervised
training and validation sets.  The same constants standardize every fit
and the evaluation inputs and outputs.  The predictors have no additional
fitted intercept.

For each of five prespecified seeds, we permute the remaining training
rows.  The first $7\times256$ rows form seven disjoint training pools;
the next $7\times256$ form seven disjoint validation pools.  Task $t$
uses its own pool and only torque $t$ as its response.  The $n=64$
condition uses the first 64 rows of each task's training pool; the
$n=256$ condition uses the entire pool.  Every method and feature
subspace uses these same paired observations. Preparation, fitting and
model selection use rows indexed from the official training file; selection
minimizes validation error. The official test file supplies the fixed
evaluation rows.

Exact numeric matching reveals overlap between the two files: all 4,449
evaluation inputs and 4,446 complete input--output observations also occur
in the training file. Of the 1,024 preparation observations, 118 also occur
in the evaluation file. The union of preparation and all tasks' training
and validation sets contains 332--346 evaluation observations per run at
$n=64$ and 465--473 at $n=256$. A task's own training subset contains
2--11 matches at $n=64$ and 19--34 at $n=256$; its validation subset contains
12--33. Preparation uses all seven outputs of its rows, whereas each
supervised task uses only its designated response. Thus disjoint indices
within the training file do not ensure disjoint evaluation observations.

The scores describe agreement between the fitted procedures on these
overlapping observations. Different seeds can reuse training rows; the
reported intervals measure variation across the five training draws
conditional on the common preparation, dictionary and evaluation file.
They do not assess prediction on an independent held-out sample.

\paragraph{A common finite input RKHS.}
In standardized input coordinates $z\in\mathbb R^{21}$, independently of
the observations, draw eight frequency vectors for each
$s\in\{1/2,1,2\}$, with distribution
$N(0,s^2I_{21}/21)$.  For the resulting 24 vectors $w_j$, let
\[
 b_{2j-1}(z)=\sqrt2\sin(w_j^\top z),
 \qquad b_{2j}(z)=\sqrt2\cos(w_j^\top z).
\]
Their 48-dimensional span has the inner product of
$L^2(\mu)$ for $\mu=N(0,I_{21})$.  Full support of $\mu$ and continuity
of the features ensure that a continuous function in this span with
zero $L^2(\mu)$ norm vanishes everywhere.  Point evaluation is continuous
in this finite-dimensional space, which is therefore an RKHS.  The
frequency vectors are nonzero and distinct up to sign almost surely,
so their sine and cosine functions are linearly independent almost
surely.  Every candidate representation selects the 16 functions in one
frequency block from this fixed ambient space.

Writing $a_{jk}=\exp(-\|w_j-w_k\|_2^2/2)$ and
$c_{jk}=\exp(-\|w_j+w_k\|_2^2/2)$, the Gaussian characteristic function
gives
\[
 Q_{2j-1,2k-1}=a_{jk}-c_{jk},\qquad
 Q_{2j,2k}=a_{jk}+c_{jk},\qquad
 Q_{2j-1,2k}=Q_{2j,2k-1}=0.
\]
The Gram matrix of all 48 features has smallest and largest eigenvalues
$0.0229770$ and $19.9595$, and condition number $868.67$.
The three block condition numbers are $551.48$, $37.91$ and $1.65$.
The implementation adds no ridge, jitter or eigenvalue truncation; it
raises an error if a matrix to be inverted has smallest eigenvalue at
most $10^{-12}$.

For each seed we independently draw 2,048 points from $\mu$ and compute
$\widehat Q=m^{-1}\sum_j b(U_j)b(U_j)^\top$ once for the full dictionary.
All empirical block Grams are submatrices of this same estimate.  Thus a
spectral approximation for the full dictionary applies simultaneously to the three subspaces and
all fitted heads, including their validation-based selection.  It does
not assert that a validation-selected subspace minimizes the training
objective jointly over all subspaces.  The Gaussian reference
law defines the function inner product independently of the SARCOS
input distribution.

\paragraph{Fitting and selection.}
For each block let $X_t$ contain its 16 features on task $t$'s training
inputs.  For $M=Q$, $M=\widehat Q$, or $M=I$ restricted to that block,
we minimize
\[
 \frac{1}{2\Task n}\sum_{t=1}^{\Task}
 \|X_tAe_t-y_t\|_2^2
 +\lambda\|M^{1/2}A/\sqrt n\|_*,\qquad \Task=7.
\]
The separated representation operator norm is a fixed scalar within a
block and is absorbed into its independently tuned $\lambda$.
With $B=M^{1/2}A/\sqrt n$ and
$Z_t=\sqrt n X_tM^{-1/2}$, accelerated proximal gradient uses singular
value thresholding for the nuclear penalty.  For each block and method,
the 17-point grid runs geometrically from $\lambda_{\max}$ to
$10^{-4}\lambda_{\max}$, where $\lambda_{\max}$ is the operator norm of
the loss gradient at zero in $B$ coordinates.  Each method selects the
lowest mean standardized validation MSE among all $3\times17$ candidates.
Exact ties favor the earlier bandwidth and then the larger penalty.
There is no refit after selection.

All 1,530 candidate fits satisfy the relative primal--dual gap tolerance
$10^{-7}$; the largest recorded gap is $9.997\times10^{-8}$.
The full paths, objective values and residual-based dual certificates
are retained.  None of the 30 selected models lies at either end of its
penalty grid.  The three methods choose identical bandwidths: for
seeds in their declared order, $(1,1/2,1,1,1/2)$ at both sample sizes.

\paragraph{Metrics and observed approximation.}
The primary metric is the arithmetic mean of seven standardized MSEs on
the overlapping evaluation observations described above. The supplementary
records also contain the seven individual standardized and original-unit
errors. The fixed preparation-mean predictor has average standardized
MSE $0.97483$ on these same rows.
For each sample size, the descriptive two-sided 95\% interval for a
paired difference uses the five-seed mean plus or minus
$t_{4,0.975}$ times its standard error.  All prescribed method comparisons
are reported, without selecting favorable seeds or task subsets.
The estimated-minus-exact mean differences are $-0.000309$ at $n=64$
and $0.000033$ at $n=256$, with intervals
$[-0.000709,0.000092]$ and $[-0.000242,0.000307]$, respectively.

The observed relative spectral errors for the full dictionary are
$0.3148$, $0.3302$, $0.3701$, $0.3611$ and $0.3335$.
Across all five realized matrices, the extremal eigenvalues yield the
deterministic uniform factors $0.84325$ and $1.17051$ for the estimated
penalty relative to the exact penalty.  These are observed matrix
comparisons.  A simple uniform leverage bound is
$L=48/\lambda_{\min}(Q)\simeq2089.04$, because
$\|b(z)\|_2^2=48$ for every $z$.
For $\varepsilon=1/2$ and $\delta=0.05$, the sufficient sample size in
\Cref{eq:feature-reference-size} with this bound is 189,521.
The actual reference sample is smaller; the application does not claim
that this conservative sufficient confidence condition is met.

\paragraph{Computation and reproducibility.}
The recorded candidate times total 7.49 seconds and include fitting,
coefficient conversion and validation evaluation. They exclude file I/O,
preparation and final selection bookkeeping. Historical thread settings
were not recorded.  A separate post-fit microbenchmark, using
100 repetitions after one warmup, gives median times of 0.067\,ms for
the analytic $48\times48$ Gram, 1.733\,ms for the empirical Gram including
one evaluation of the 2,048-point feature matrix, and 0.208\,ms when that
feature matrix is already available.  An analytic Gram is inexpensive
in this example; these timings describe implementation costs rather
than a computational advantage for estimation.

The machine-readable protocol and solver source were hashed and locked
before any predictive fit.  The supplement retains the source, official
download URLs and SHA-256 hashes, preparation and supervised split
indices, preprocessing constants, frequencies, reference points, Grams,
all fitted coefficient paths and all metrics.  The official raw
observations are downloaded and verified by the script and are excluded
from redistribution because the hosting page specifies no dataset
license.  This experiment is a prespecified exploratory application.

\FloatBarrier

\section{Task-count regression studies and reproducibility}\label{app:legacy-experiments}

We compare task-count weighting and penalty choice in controlled linear multitask regression. The two suites vary rank, sample imbalance and the relative orientation of shared and task factors. They complement the exact composition examples in \Cref{sec:no-go} by showing how the choice of penalty affects prediction error in a setting where population excess can be evaluated directly.

\subsection{Design and estimator}\label{sec:locked-protocol}

For task designs $X_t$ and responses $y_t$, let
\begin{align}
\widehat L(F)
=
\frac{1}{2\Task}
\sum_{t=1}^{\Task}
\frac{1}{n_t}
\left\lVert X_tFe_t-y_t\right\rVert_2^2.
\label{eq:experimental-loss}
\end{align}

For a positive diagonal count matrix $M$, the nuclear estimators take the form
\begin{align}
\widehat F_M(\lambda)\in\arg\min_F
\left\{\widehat L(F)+\lambda\lVert FM^{-1/2}\rVert_*\right\},
\label{eq:intro-direct-estimator}
\end{align}

The weighted-joint estimator minimizes
\begin{align}
\widehat L(F)
+
\lambda
\left\lVert F\Dn^{-1/2}\right\rVert_*.
\label{eq:experimental-estimator}
\end{align}
The unweighted and shifted-count nuclear controls replace $\Dn$ by $I$ and $\operatorname{diag}(\operatorname{roll}(\bn,1))$. Weighted Frobenius and independent ridge use $\lambda\lVert F\Dn^{-1/2}\rVert_{\HS}^2/2$ and $\lambda\lVert F\rVert_{\HS}^2/2$, respectively. Ridge task systems are independent conditional on a single jointly selected $\lambda$. Each method selects from its own normalized path by minimizing $0.5$ times task-average validation MSE plus $0.5$ times low-resource validation MSE. Nuclear paths use the zero-solution scale $\lambda_{\max}=\lVert\nabla_B\widehat L(0)\rVert_{\opn}$ for $B=FM^{-1/2}$; implementation details are in \Cref{app:direct-implementation}.

The design rows are independent $N(0,I_d)$ and $y_{ti}=x_{ti}^{\top}f_t^\star+\epsilon_{ti}$, with independent $N(0,\sigma^2)$ noise. \Cref{tab:setup-metrics} gives the sample sizes and metrics. Each scenario uses independent training and validation observations to fit and select its own model. The development/held-out angle labels specify which scenarios enter the final summaries; tuning is performed within each scenario.

\begin{table}[!htbp]
\centering\small
\setlength{\tabcolsep}{3pt}
\renewcommand{\arraystretch}{1.15}
\begin{tabulary}{\linewidth}{@{}lYY@{}}
\toprule
Quantity & Replication & Structural transfer \\
\midrule
Input dimension / tasks & $8/12$ & $10/15$ \\
Ranks & $1,2,4$ & $1,3,5$ \\
Imbalance ratios $\rho$ & $1,8,32$ & $1,4,16,64$ \\
Minimum train count & $24$ & $20$ \\
Validation / test per task & $64/256$ & $80/320$ \\
Noise standard deviation & $0.75$ & $0.8$ \\
Low-resource tasks & $3$ & $3$ \\
\bottomrule
\end{tabulary}
\caption{Data and metric setup. Training counts are nearest integers on a geometric progression from the minimum count to $\rho$ times that count, with exact endpoints. Population excess MSE is $\lVert F-F^\star\rVert_{\HS}^2/\Task$, without the optimization loss's factor $1/2$. Low-resource excess averages squared column errors over the first $\max(1,\lfloor\Task/4\rfloor)$ indices obtained by sorting the counts; the implementation uses NumPy's default sorting order for ties.}
\label{tab:setup-metrics}
\end{table}

The two suites contain $108+144=252$ paired evaluation scenarios. \Cref{app:truth-pairing} specifies the pairing and the suite-level and pooled resampling schemes.

\subsubsection{Truth family, scenario labels and pairing}\label{app:truth-pairing}

The generating truth is
\begin{align*}
F^\star(\theta)&=KQ(\theta)_{:,1:r}S_rV_{:,1:r}^{\top},
&Q(\theta)&=\exp(\theta A)Q_0.
\end{align*}
Here $K$ and $S_r$ are diagonal with geometrically spaced entries, $Q_0$ and $V$ are orthogonal matrices from Gaussian QR factorizations, and $A=(H-H^\top)/\lVert H-H^\top\rVert_{\opn}$ for a Gaussian matrix $H$. For rank one, the single task singular value is its maximum endpoint. Bases and generator are fixed within each suite. \Cref{tab:appendix-suite-config} gives their parameters.

\begin{table}[!htbp]
\centering\small
\setlength{\tabcolsep}{3pt}
\renewcommand{\arraystretch}{1.15}
\begin{tabulary}{\linewidth}{@{}lYY@{}}
\toprule
Parameter & Replication & Transfer \\
\midrule
Shared spectral endpoints & $4,\ 0.25$ & $3.5,\ 0.2$ \\
Task spectral endpoints & $1.5,\ 0.4$ & $1.4,\ 0.35$ \\
Basis / generator / right-basis seeds & $117/223/329$ & $811/823/829$ \\
Global seed & $20261117$ & $20261203$ \\
Data seeds & $101,102,103$ & $201,202,203$ \\
Evaluated angles & $0.3,0.7,1.1,1.5$ & $0.25,0.65,1.05,1.45$ \\
Development-labelled angles & $0.1,0.5,0.9,1.3,1.7$ & $0.05,0.45,0.85,1.25,1.65$ \\
Evaluated pairs / groups & $108/27$ & $144/36$ \\
\bottomrule
\end{tabulary}
\caption{Fixed geometry and scenario parameters in the supplied suite configurations. Dimensions, counts and metric definitions appear in \Cref{tab:setup-metrics}. Angles are parameters of the orthogonal path $\exp(\theta A)Q_0$.}
\label{tab:appendix-suite-config}
\end{table}

Each angle selects its own regularization parameter on independent validation observations. Only scenarios labelled held-out enter the reported effects; development-labelled scenarios are excluded from these summaries. Within each suite--rank--imbalance--data-seed--split group, designs and noise are shared across methods and angles. Distinct groups and splits use distinct pseudorandom streams indexed by the global seed, rank, imbalance, data seed and split code. This pairing isolates method differences without treating the four angles within a group as independent data replications.

Effects are baseline excess minus weighted-joint excess. The suite bootstraps use $2{,}000$ replicates each, with seeds $20261201$ for replication and $20261219$ for transfer. Each replicate samples complete rank--imbalance--data-seed groups and retains all four evaluated angles. The pooled analysis applies the same scheme to all $63$ suite-labelled groups, using $5{,}000$ replicates and seed $20270317$. The interval endpoints are the $2.5\%$ and $97.5\%$ quantiles. These intervals describe resampling variation across a designed scenario collection; they are not confidence intervals for a randomly sampled population of architectures. The Gaussian squared-loss experiments are separate from the bounded-loss guarantee in \Cref{thm:population}.

\subsubsection{Direct convex implementation}\label{app:direct-implementation}

For a method-specific diagonal count matrix $M$, the code solves \Cref{eq:experimental-estimator} in the transformed variable $B=FM^{-1/2}$.  The smooth gradient has independent task blocks, and the nuclear proximal map is singular-value soft thresholding.  The FISTA step is $0.99/L_M$, where $L_M$ is the largest transformed task Hessian eigenvalue.  Adaptive restart is used, and the selected solution must satisfy the proximal fixed-point residual
\begin{align*}
\frac{1}{\eta}
\left\lVert
B-
\operatorname{SVT}_{\eta\lambda}
\left(B-\eta\nabla\widehat L(B)\right)
\right\rVert_{\HS}
\end{align*}
below $\max(10^{-8},10\times10^{-10})=10^{-8}$. The configured iterate tolerance is $10^{-10}$; the separate acceptance gate permits selected nuclear residuals up to $10^{-6}$.  Weighted Frobenius and independent ridge are solved by their exact taskwise linear systems.  The zero-solution threshold described in \Cref{sec:locked-protocol} is evaluated at every nuclear path endpoint by the supplied runner.

The method-specific nuclear fractions are
\begin{align*}
1.5,
1,
0.5,
0.2,
0.1,
0.05,
0.02,
0.01,
0.003,
0.001,
0.0003,
\end{align*}
and the smooth-control path spans fractions from $300$ to $10^{-4}$ of its transformed Hessian scale.  Each scenario selects its path fraction from validation data only, using the common criterion in \Cref{sec:locked-protocol}. The supplementary material includes the solver tests and the original model and path diagnostics. The numerical summaries and archive scope are described in \Cref{app:release-status}.

\FloatBarrier
\subsection{Results}\label{sec:confirmatory-results}

\begin{figure}[!htbp]
\centering
\begin{minipage}[t]{0.49\linewidth}
\centering\vspace{0pt}\textbf{(a)}\par\smallskip
\includegraphics[width=\linewidth]{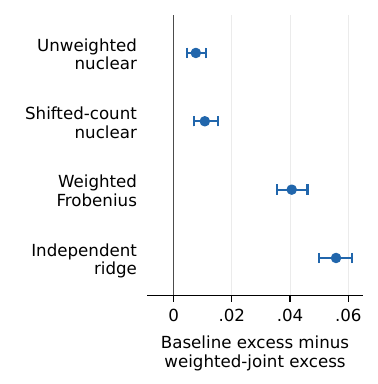}
\end{minipage}\hfill
\begin{minipage}[t]{0.49\linewidth}
\centering\vspace{0pt}\textbf{(b)}\par\smallskip
\includegraphics[width=\linewidth]{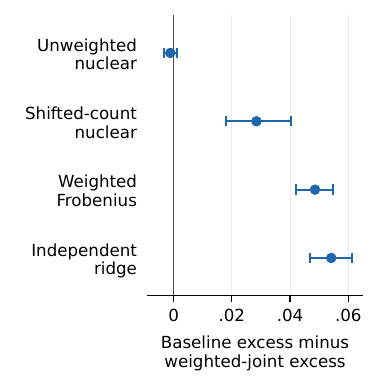}
\end{minipage}
\caption{Paired effects over $252$ evaluation scenarios. Points show mean effects and bars show $95\%$ grouped-bootstrap intervals; positive values favor weighted joint nuclear. (a) Population excess; (b) least-sampled-quartile excess}
\label{fig:phase2-effects}
\end{figure}

We report paired effects as baseline excess minus weighted-joint excess, so positive values favor the weighted nuclear penalty. Against unweighted nuclear regularization, weighted joint nuclear reduces average population excess by $0.00764$, with $95\%$ grouped-bootstrap interval $[0.00465,0.01110]$. Both suite means are positive; \Cref{tab:suite-effects} gives the separate estimates. Figure~\ref{fig:phase2-effects} summarizes the pooled effects for both metrics.

\begin{figure}[!htbp]
\centering
\begin{minipage}[t]{0.49\linewidth}
\centering\vspace{0pt}\textbf{(a)}\par\smallskip
\includegraphics[width=\linewidth]{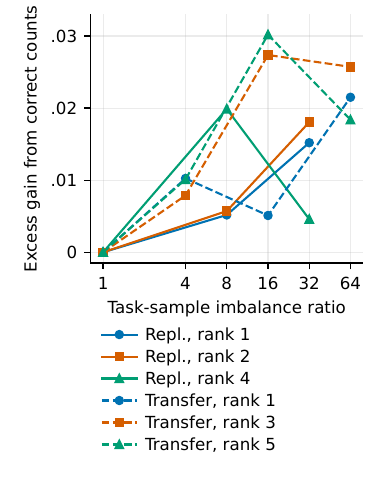}
\end{minipage}\hfill
\begin{minipage}[t]{0.49\linewidth}
\centering\vspace{0pt}\textbf{(b)}\par\smallskip
\includegraphics[width=\linewidth]{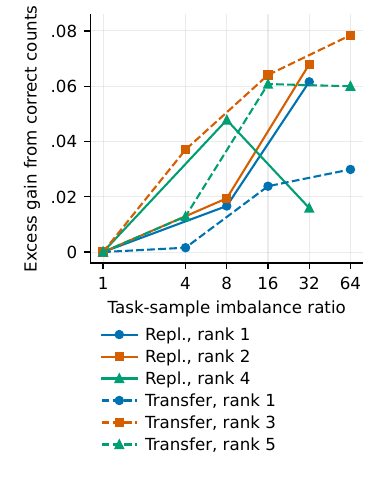}
\end{minipage}
\caption{Correct-count gains by suite, rank, and imbalance.  Effects are shifted-count excess minus correct-count excess.  They vanish at balance and are positive in all $15$ imbalanced cells for both metrics.  (a) Population excess; (b) least-sampled-quartile excess}
\label{fig:count-cells}
\end{figure}

Correct counts improve population and least-sampled-quartile excess over shifted counts by $0.01072$ $[0.00686,0.01531]$ and $0.02847$ $[0.01805,0.04045]$.  Both suite lower bounds are positive, all $15$ imbalanced cells have positive means for both metrics, and balanced cells have exactly zero effect.  Figure~\ref{fig:count-cells} shows the count-weighting effects across imbalance levels.

The population and low-resource effects are $0.04057/0.04857$ against weighted Frobenius and $0.05580/0.05415$ against independent ridge, with positive intervals. The low-resource comparison with unweighted nuclear varies across suites: the pooled effect is $-0.00112$ $[-0.00336,0.00124]$; the suite estimates in \Cref{tab:suite-effects} have opposite signs. Correct count weighting improves the task-average error in these suites, while the least-sampled tasks reveal a tradeoff between the two nuclear penalties.

\Needspace{4\baselineskip}
\subsubsection{Suite-level effects}

Table~\ref{tab:suite-effects} reports the suite-level means and $95\%$ grouped-bootstrap intervals.

\begin{table}[!htbp]
\centering
\small
\setlength{\tabcolsep}{2.8pt}
\renewcommand{\arraystretch}{1.15}
\begin{tabulary}{\textwidth}{@{}llYYc@{}}
\toprule
Suite & Baseline & Population effect & Low-resource effect & Pairs \\
\midrule
Replication & Unweighted nuclear
& \shortstack[l]{$0.00217$\\$[0.00046,0.00434]$}
& \shortstack[l]{$-0.00502$\\$[-0.00755,-0.00258]$} & $108$ \\
Transfer & Unweighted nuclear
& \shortstack[l]{$0.01174$\\$[0.00714,0.01717]$}
& \shortstack[l]{$0.00181$\\$[-0.00145,0.00516]$} & $144$ \\
Replication & Shifted counts
& \shortstack[l]{$0.00762$\\$[0.00330,0.01302]$}
& \shortstack[l]{$0.02548$\\$[0.01206,0.04158]$} & $108$ \\
Transfer & Shifted counts
& \shortstack[l]{$0.01304$\\$[0.00718,0.01986]$}
& \shortstack[l]{$0.03071$\\$[0.01568,0.04827]$} & $144$ \\
Replication & Weighted Frobenius
& \shortstack[l]{$0.04059$\\$[0.03163,0.05071]$}
& \shortstack[l]{$0.06266$\\$[0.05397,0.07025]$} & $108$ \\
Transfer & Weighted Frobenius
& \shortstack[l]{$0.04056$\\$[0.03504,0.04658]$}
& \shortstack[l]{$0.03800$\\$[0.03036,0.04556]$} & $144$ \\
Replication & Independent ridge
& \shortstack[l]{$0.04671$\\$[0.03807,0.05622]$}
& \shortstack[l]{$0.06857$\\$[0.05872,0.07706]$} & $108$ \\
Transfer & Independent ridge
& \shortstack[l]{$0.06261$\\$[0.05639,0.06931]$}
& \shortstack[l]{$0.04332$\\$[0.03476,0.05235]$} & $144$ \\
\bottomrule
\end{tabulary}
\caption{Suite-level paired effects.  Positive values favor weighted joint nuclear regularization.}
\label{tab:suite-effects}
\end{table}

\subsection{Numerical checks and archived records}\label{app:release-status}\label{app:reproducibility}

The protocol specifies six effect criteria and a numerical criterion covering both suites. \Cref{tab:gate-evidence} gives the thresholds and the corresponding results. The supplementary archive contains the original configuration snapshots, selected-model records, regularization paths, paired effects, pooled bootstrap draws and run diagnostics. It also includes the protocol manifest and the original output checksums.

The reduced supplementary archive contains \path{frozen_results/phase2_pooled_effects.csv}, \path{frozen_results/phase2_suite_effects.csv} and \path{frozen_results/phase2_count_weighting_cells.csv}. The accompanying original-record archive provides the underlying replication, transfer and pooled output directories. Pooled means agree with the pair-count-weighted suite means, and count-cell means agree with their corresponding suite means.

The vector figures are generated with Matplotlib by the accompanying \path{generate_revised_figures.py}. Figures~\ref{fig:phase2-effects} and \ref{fig:count-cells} use the pooled-effect and count-cell CSV files in the supplementary archive.

\begin{table}[!tbp]
\centering\small
\setlength{\tabcolsep}{3pt}
\renewcommand{\arraystretch}{1.15}
\begin{tabulary}{\linewidth}{@{}lYY@{}}
\toprule
Criterion & Threshold & Result \\
\midrule
Unweighted population & Mean $\ge0.001$; pooled lower endpoint and suite means positive & $0.00764$; lower $0.00465$; both means positive \\
Shifted-count population & Mean $\ge0.004$; pooled/suite lower endpoints positive; $\ge13$ positive cells & $0.01072$; lower $0.00686$; $15/15$ cells \\
Shifted-count low resource & Mean $\ge0.010$; analogous interval and cell criteria & $0.02847$; lower $0.01805$; $15/15$ cells \\
Weighted Frobenius & Both means $\ge0.020$ and lower endpoints positive & $0.04057/0.04857$; both intervals positive \\
Independent ridge & Same as Frobenius & $0.05580/0.05415$; both intervals positive \\
Balanced-count null & Maximum absolute cell mean $\le10^{-12}$ & $0$ \\
Numerical/path checks & Both suite runners satisfy the configured checks & Original run diagnostics satisfy all thresholds; archive scope is described below \\
\bottomrule
\end{tabulary}
\caption{Protocol criteria and experimental results. Numerical diagnostics are detailed below.}
\label{tab:gate-evidence}
\end{table}

\looseness=-1
The configured numerical checks require marginal-spectrum error and relative separated-norm spread at most $10^{-12}$, maximum joint-norm range ratio at least $1.05$, finite model and path records, zero-endpoint norm at most $10^{-8}$, selected nuclear and ridge residuals at most $10^{-6}$ and $10^{-10}$, minimum method convergence rate $0.99$, maximum path-edge selection rate $0.10$, and matching random draws across paired scenarios. The model and path records contain $1{,}215/1{,}620$ selected fits and $14{,}823/19{,}764$ path candidates in replication/transfer. All selected fits are marked as converged. The largest stored nuclear residuals are $9.72\times10^{-9}$ and $9.94\times10^{-9}$; the largest stored ridge residuals are $7.02\times10^{-17}$ and $6.05\times10^{-17}$. Maximum method-level path-edge selection rates are $1.65\%$ and $0\%$.

The archived paired effects reproduce the reported means, and the saved pooled bootstrap draws reproduce the pooled interval endpoints. Original training arrays, fitted coefficient matrices and generating-factor arrays were not serialized. Consequently, the archived residuals can be aggregated from the saved diagnostics, while recalculating them from fitted models would require rerunning the fits. The common-random-number and marginal-spectrum checks are recorded in the original run summaries.

The original analysis counts a win whenever a paired effect is strictly positive. CSV parsing rounds ten positive effects with magnitude below $10^{-16}$ to zero between the suite and pooled records, accounting for their different win counts. We therefore report mean effects and intervals, whose displayed values are unaffected.

\end{document}